%% file: main.tex
\documentclass[notitlepage,12pt]{jedm}

\usepackage[table]{xcolor}
\usepackage{url}
\usepackage{hyperref}

\include{template/packages}

\include{template/commands}

\include{template/formatting}
\DeclareRobustCommand{\zhaangambiguate}[3]{#2~#3}

\hypersetup{
  colorlinks   = true, %
  urlcolor     = blue, %
  linkcolor    = blue, %
  citecolor   = blue %
}

\newcommand{\revinsert}[1]{#1}
\newcommand{\revremove}[1]{}

\begin{document}

\title{Assessing the Performance of Online \\ Students - New Data, New Approaches, Improved Accuracy}

\date{} %

\author{
{\large Robin Schmucker}\\
Carnegie Mellon University\\
rschmuck@cs.cmu.edu
\and
{\large Jingbo Wang}\\
Carnegie Mellon University\\
jingbow@andrew.cmu.edu \hspace{2.53cm} \hphantom{1px}\\
\and
{\large Shijia Hu}\\
Carnegie Mellon University\\
shijiah@andrew.cmu.edu \\
\and
{\large Tom M. Mitchell}\\
Carnegie Mellon University\\
tom.mitchell@cs.cmu.edu\\
}

\maketitle

\begin{abstract}
We consider the problem of assessing the changing \revinsert{performance levels}\revremove{knowledge state} of individual students as they go through online courses. This student performance modeling problem\revremove{, also known as knowledge tracing,} is a critical step for building adaptive online teaching systems. Specifically, we conduct a study of how to utilize various types and large amounts of log data from earlier students to train accurate machine learning models that predict the \revinsert{performance}\revremove{knowledge state} of future students. This study is the first to use four very large sets of student data made available recently from four distinct intelligent tutoring systems.

Our results include a new machine learning approach that defines a new state of the art for \revinsert{logistic regression based} student performance modeling, improving over earlier methods in several ways: First, we achieve improved accuracy of student modeling by introducing new features that can be easily computed from conventional question-response logs (e.g., features such as the pattern in the student's most recent answers). Second, we take advantage of features of the student history that go beyond question-response pairs (e.g., features such as which video segments the student watched, or skipped) as well as background information about prerequisite structure in the curriculum. Third, we train multiple specialized student performance models for different aspects of the curriculum (e.g., specializing in early versus later segments of the student history), then combine these specialized models to create a group prediction of the student \revinsert{performance}\revremove{knowledge state}. Taken together, these innovations yield an average AUC score across these four datasets of 0.808 compared to the previous best logistic regression approach score of 0.767, and also outperforming state-of-the-art deep neural net approaches. Importantly, we observe consistent improvements from each of our three methodological innovations, in each diverse dataset, suggesting that our methods are of general utility and likely to produce improvements for other online tutoring systems as well.

{\parindent0pt
\textbf{Keywords:} performance modeling, knowledge tracing, logistic regression, deep learning, features
}
\end{abstract}

\input{text/introduction}
\input{text/related_work}

\input{text/datasets}
\input{text/approach}
\input{text/experiments}
\input{text/discussion}

\input{text/conclusion}

\appendix
\input{text/appendix}

\section*{Acknowledgments}

We would like to thank Squirrel Ai Learning for providing the \texttt{Squirrel Ai ElemMath2021} dataset. In particular, Richard Tong and Dan Bindman were instrumental in providing this dataset, in helping us understand it.  
We thank Jack Wang and Angus Lamb, for personal communication enabling us to reconstruct the exact sequence of student interactions in the \texttt{Eedi} data logs. We thank Theophile Gervet for making his earlier implementations of several algorithms available in his code repository, and for help understanding the details. We are grateful to the teams at the CK-12 Foundation and at Squirrel Ai Learning for suggestions on how to improve the presentation in this paper. This work was supported in part through the CMU - Squirrel Ai Research Lab on Personalized Education at Scale, and in part by AFOSR under award FA95501710218.

\bibliographystyle{acmtrans}
\DeclareRobustCommand{\zhaangambiguate}[3]{#1}
\bibliography{./bibliography}

\end{document}

%% file: template/packages.tex
\usepackage{booktabs}       %
\usepackage{amsfonts}       %
\usepackage{amsmath}
\usepackage{amssymb}
\usepackage{amsthm}
\usepackage{mathtools}
\usepackage{nicefrac}       %
\usepackage{microtype}      %
\usepackage{enumitem}
\usepackage{dsfont}
\usepackage{hyperref}
\usepackage{siunitx}
\usepackage{bm}
\usepackage{xcolor}
\usepackage{tcolorbox}
\usepackage{mleftright}
\usepackage{stmaryrd}
\usepackage{numprint}
\usepackage{graphicx}
\usepackage{float}
\usepackage{pifont}
\usepackage{cleveref}
\usepackage{subcaption}
\usepackage[linesnumbered,ruled,lined,noend]{algorithm2e}
\usepackage{wrapfig}
\usepackage[]{minitoc}
\usepackage{thm-restate}
\usepackage{tabularx}
\usepackage{multirow}
\usepackage{array}
\usepackage{rotating}

%% file: template/commands.tex
\newcommand{\yes}{\ding{51}}
\newcommand{\no}{\ding{55}}
\newcommand{\noo}{}
\renewcommand{\vec}[1]{\bm{#1}}

\newcommand{\trans}{^{\!\top}}

\usepackage[nomessages]{fp}
\newcommand{\avgvar}[2]{\FPset\x{#1} \FPmul{\x}{\x}{100} \FPeval{\x}{round(\x:2)} \FPprint\x}

\newcolumntype{C}{>{\raggedright\arraybackslash}X}

%% file: template/formatting.tex
\DeclareMathSizes{10}{9}{7}{5}
\makeatletter
\g@addto@macro \normalsize {%
 \addtolength\abovedisplayskip{-3pt}%
 \addtolength\belowdisplayskip{-3pt}%
}
\makeatother

%% file: text/introduction.tex
\section{Introduction}
\label{sec:introduction}

Intelligent online tutoring systems (ITS's) are now used by millions of students worldwide, enabling access to quality teaching materials and to personally customized instruction. These systems depend critically on their ability to track the evolving \revinsert{ability level}\revremove{knowledge state} of the student, in order to deliver the most effective instructional material at each point in time. Because the problem of assessing the student's evolving \revinsert{ability to solve different questions}\revremove{knowledge state} (also referred to as student performance modeling\revremove{, or knowledge tracing (KT)}) is so central to successfully customizing instruction to the individual student, it has received significant attention in recent years.  

The state of the art that has emerged for this student performance modeling problem involves applying machine learning algorithms to historical student log data. The result produced by the machine learning algorithm is a model, or computer program, that outputs the estimated \revinsert{likelihood of correct response}\revremove{knowledge state} of any future student \revinsert{for any particular question} at any point in the lesson, given the sequence of steps they have taken up to this point in the lesson (e.g.,~\citeNP{Corbett1994:Knowledge}, \citeNP{Pavlik2009:Performance}, \citeNP{Piech2015:Deep}, \citeNP{Pandey2019:Self}, \citeNP{Gervet2020:Deep}, \citeNP{Shin2021:Saint+}). These systems typically represent the student knowledge state as a list of probabilities that the student will correctly answer a particular list of questions that cover the key concepts (also known as "knowledge components" (KC's)) in the curriculum. Most current approaches estimate the student's\revremove{knowledge} state by considering only the log of questions asked and the student's answers, though recent datasets provide considerably more information such as the length of time taken by the student to provide answers, specific videos the student watched and whether they watched the entire video, and what hints they were given as they worked through specific practice problems.

This paper seeks to answer the question of which machine learning approach produces the most accurate estimates of students' \revinsert{ability to solve different questions}\revremove{knowledge states}. To answer this question we perform an empirical study using data from over 750,000 students taking a variety of courses, to study several aspects of the question including (1) which types of machine learning algorithms work best? (2) which features of a student's previous and current interactions with the ITS are most useful for predicting their current \revinsert{ability to solve a certain question}\revremove{knowledge state}? (3) how valuable is background information about curriculum prerequisites for improving accuracy? and (4) can accuracy be improved by training specialized models for different portions of the curriculum? We measure the quality of alternative approaches by how accurately they predict which future questions the student answers correctly.

More specifically, we present here the first comparative analysis of recent state-of-the-art algorithms for student performance modeling across four very large student log datasets that have recently become available, which are each approximately 10 times larger than earlier publicly available datasets, which cover a variety of courses in elementary mathematics, as well as teaching English as a second language, and which range across different teaching objectives such as initial assessment of student knowledge state, test preparation, and extra-curricular tutoring complementing K-12 schooling. We show that accuracy of student performance modeling can be improved beyond the current state of the art through a combination of techniques including incorporating new features from student logs (e.g., time spent on previously answered questions), incorporating background information about prerequisite/postrequisite topics in the curriculum, and training multiple specialized models for different parts of the student experience (e.g., training distinct models to assess new students during the first 10 steps of their lesson, versus students taking the post-lesson quiz). The fact that we see consistent improvements in accuracy across all four datasets suggests the lessons gained from our experiments are fairly general, and not tied to a specific type of course or specific tutoring system.

To summarize, the key contributions of this paper include:

\begin{itemize}

\item {\em Cross-ITS study on modern datasets.} We present the first comparative analysis of state-of-the-art approaches to student performance modeling across four recently published, large and diverse student log datasets taken from four distinct intelligent tutoring systems (ITS's), resulting in the largest empirical study to date of student performance modeling. These four systems teach various topics in elementary mathematics or English as a second language, and the combined data covers approximately 200,000,000 observed actions taken by approximately 750,000 students. Three of these datasets have been made publicly available over the past few years. %

\item {\em Improved student performance modeling by incorporating prerequisite and hierarchical structure across knowledge components.} All four of the datasets we consider provide a vocabulary of knowledge components (KC's) taught by the system. Two datasets provide meta-information about which KCs are prerequisites for which others (e.g., {\em Add and subtract negative numbers} is a prerequisite for {\em Multiply and divide negative numbers}), and one provides a hierarchical structure (e.g., the KC {\em Add and subtract vectors} falls hierarchically under {\em Basic vectors}). We found that incorporating the pre- and postrequisite information into the model resulted in substantial improvements in accuracy predicting which future questions the student would answer correctly (e.g., by including features such as the counts of correctly answered questions related to prerequisite and post-requisite topics). Features derived from the hierarchical structure lead to improvements as well, but to a lesser degree than the ones extracted from the prerequisite structures.

\item {\em Improved student performance modeling by incorporating log data features beyond which questions were answered correctly.} Although earlier work has focused on student performance models that consider only the sequence of questions asked and which were answered correctly, modern log data includes much more information. We found that incorporating this information yields accuracy improvements over the current state of the art. For example, we found that features such as the length of time the student took to answer previous questions, the number of videos watched on the KC, and whether the student is currently answering a question in a pre-test, post-test, or a practice session were all useful features for predicting whether a student would correctly answer the next question.

\item {\em Improved student performance modeling by training multiple models for specialized contexts, then combining them.} We introduce a new approach of training multiple distinct student assessment models for distinct learning contexts. For example, training a distinct model for the "cold start" problem of assessing students who have just begun the course and have little log data at this point, yields significant improvements in the accuracy of student \revinsert{performance predictions}\revremove{knowledge state assessment}.  Furthermore, combining predictions of multiple specialized models (e.g., one trained for the current study module type, and one trained for students that have answered at least $n$ questions in the course) leads to even further improvements. 

\item {\em The above improvements can be combined.} The above results show improvements in student performance modeling due to multiple innovations. By combining these we achieve overall improvements over state-of-the-art logistic regression methods that reduce AUC error by $17\revinsert{.5}\%$ on average over these four datasets, as summarized in Table~\ref{tab:summaryResults}.

\input{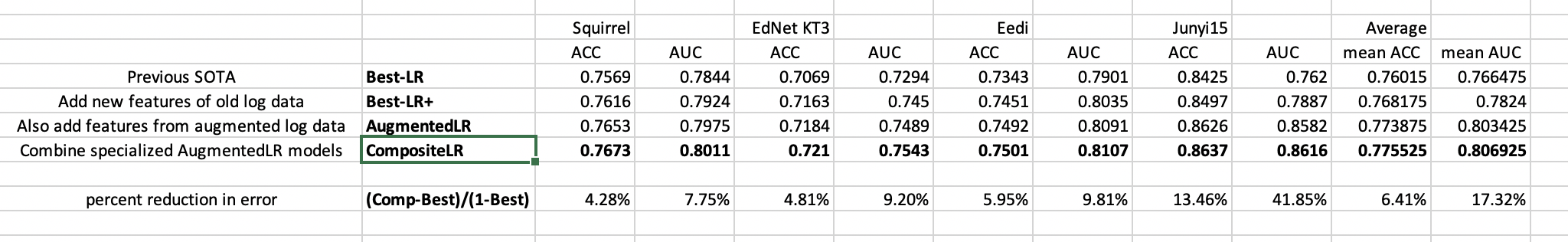}

\end{itemize}

%% file: tables/LRsummaryTable.tex
\begin{table}[t]
\scriptsize
\centering
\caption{Improvements to state of the art (SOTA) in student performance modeling, due to innovations introduced in this paper.  The previous logistic regression state-of-the-art approach is Best-LR.  Performance across each of the four diverse datasets improves with each of our three suggested extensions to the Best-LR algorithm.  Best-LR+ extends Best-LR by adding new features calculated from the question-response (Q-R) data available in most student logs. AugmentedLR further adds a variety of novel features that go beyond question-response data (e.g., which videos the student watched or skipped). Combined-AugmentedLR further extends the approach by training multiple logistic regression models (Multimodel) on different subsets of the training data (e.g., based on how far into the course the student is currently). Together, these extensions to the previous state of the art produce substantial improvements across each of these four datasets, improving the average AUC score from 0.767 to 0.808 on average across the four datasets -- a reduction of \revinsert{17.5\%}\revremove{17.3\%} in average AUC error (i.e., in the difference between the observed AUC and the ideal perfect AUC of 1.0).}
\begin{tabular}{|l|cc|cc|cc|cc|cc|}
\hline
& \multicolumn{2}{c|}{\texttt{ElemMath2021}} & \multicolumn{2}{c|}{\texttt{EdNet KT3}} & \multicolumn{2}{c|}{\texttt{Eedi}} & \multicolumn{2}{c|}{\texttt{Junyi15}} & \multicolumn{2}{c|}{Average} \\
    & ACC & AUC & ACC & AUC & ACC & AUC & ACC & AUC & ACC & AUC \\
\hline
Prev. SOTA: Best-LR  & 0.7569 & 0.7844 & 0.7069 & 0.7294 & 0.7343 & 0.7901 & 0.8425 & 0.7620 & 0.7602 & 0.7665 \\
Add Q-R features: Best-LR+      & 0.7623 & 0.7935 & 0.7169 & 0.7465 & 0.7455 & 0.8040 & 0.8505 & 0.7912 & 0.7688 & 0.7838 \\
Add novel features: AugmLR        & 0.7659 & 0.7987 & 0.7189 & 0.7500 & 0.7496 & 0.8096 & 0.8635 & 0.8603 & 0.7745 & 0.8047 \\
Multimodel: Comb-AugmLR      & \textbf{0.7676} & \textbf{0.8016} & \textbf{0.7211} & \textbf{0.7548} & \textbf{0.7504} & \textbf{0.8111} & \textbf{0.8646} & \textbf{0.8634} & \textbf{0.7759} & \textbf{0.8077} \\
\hline
\hline %
percent error reduction & 4.40\% & 7.98\% & 4.84\% & 9.39\% & 6.06\% & 10.00\% & 14.03\% & 42.61\% & 7.33\% & 17.50\%
 \\
\hline
\end{tabular}
\label{tab:summaryResults}
\end{table}

%% file: text/related_work.tex
\section{Related Work}
\label{sec:related work}

Student performance modeling techniques estimate a student's \revinsert{likelihood to solve different problems}\revremove{knowledge state} based on their interactions with the ITS. These performance models and the estimates of student proficiency they produce are a key component of current ITS's which allow the tutoring system to adapt to each student's personal \revinsert{ability level}\revremove{knowledge state} at each point in the curriculum. In the literature, performance modeling is also sometimes referred to as knowledge tracing, proficiency modeling, or student assessment. There are three main categories of performance modeling techniques: (i) Markov process based probabilistic modeling, (ii) logistic regression and (iii) deep learning based approaches.

Markov process based techniques, such as Bayesian Knowledge Tracing (BKT)~\cite{Corbett1994:Knowledge} and its various extensions~(\citeNP{Baker2008:More}, \citeNP{Pardos2010:Modeling}, \citeNP{Pardos2011:KT}, \citeNP{Qiu2011:Does}, \citeNP{Yudelson2013:Individualized}, \citeNP{Sao2013:Incorporating}, \citeNP{Khajah2016:Deep}, \citeNP{Kaser2017:Dynamic}) have a long history in the educational data mining (EDM) community. Most approaches in this family determine a student proficiency by performing probabilistic inference using a two state Hidden Markov Model containing one state representing that the student has \emph{mastered} a particular concept, and one state representing \emph{non-mastery}. A recent study comparing various performance modeling algorithms across nine real-world datasets~\cite{Gervet2020:Deep} found that when applied to large-scale datasets, BKT and its extension BKT+~\cite{Khajah2016:Deep} are very slow to train and their predictive performance is not competitive with more recent logistic regression and deep learning based approaches. The Python package pyBKT was released and promises faster training times for Bayesian Knowledge Tracing models~\cite{Badrinath2021:pybkt}.

Logistic regression models take as input a vector of manually specified features calculated from a student's interaction history, then output a predicted probability that this student has mastered a particular concept or KC (often implemented as the probability that they will correctly answer a specified question). Common approaches include IRT~\cite{Van2013:Handbook}, LFA~\cite{Cen2006:Learning}, PFA~\cite{Pavlik2009:Performance}, DASH~(\citeNP{Lindsey2014:Improving}, \citeNP{Gonzalez2014:General}, \citeNP{Mozer2016:Predicting}) and its extension DAS3H~\cite{Choffin2019:DAS3H} as well as Best-LR~\cite{Gervet2020:Deep}. While there exists a variety of logistic regression models, they mainly rely on two types of features: (i) One-hot encodings\footnote{A one-hot encoding of the question ID, for example, is a vector whose length is equal to the number of possible questions. The vector contains $n-1$ zeros, and a single $1$ to indicate which question ID is being encoded.} of question and KC identifiers; and (ii) Count features capturing the student's number of prior attempts to answer questions, and the number of correct and incorrect responses. \revinsert{R-PFA~\cite{Galyardt2015:Move} augments PFA with features that represent the recency-weighted count of prior incorrect responses and the recency-weighted proportion of correct responses. PPE~\cite{Walsh2018:Mechanisms} considers the timing of individual practice sessions to describe spacing effects impacting memorization in the context of learning word pairs via power functions.} DAS3H incorporates a temporal aspect into its predictions by computing count features for different time windows.
LKT~\cite{Pavlik2020:Logistic} is a flexible framework for logistic regression based student performance modeling which offers a variety of features based on question-answering behaviour. Among others it offers decay functions to capture recency effects as well as features based on the ratio of correct and incorrect responses.
Section~\ref{sec:approach} discusses multiple regression models in more detail and proposes alternative features which are able to incorporate rich information from various types of log data.

Deep learning based models, like logistic regression models, take as input the student log data, and output a predicted probability that the student will answer a specific question correctly. However, unlike logistic regression, deep learning models have the ability to automatically define useful features computable from the sequence of log data, without relying on human feature engineering. A wide range of neural architectures have been proposed for student performance modeling. DKT~\cite{Piech2015:Deep} is an early work that uses Long Short Term Memory (LSTM) networks~\cite{Hochreiter1997:Long} processing student interactions step-by-step. DKVNM~\cite{Zhang2017:Dynamic} is a memory augmented architecture that can capture multi-KC dependencies. CKT~\cite{Shen2020:Convolutional} uses a convolutional neural network~\cite{Lecun1999:Object} to model individualized learning rates. Inspired by recent advances in the natural language processing (NLP) community, multiple transformer~\cite{Ashish2017:Attention} based approaches have been proposed~(SAKT, \cite{Pandey2019:Self}; AKT, \cite{Ghosh2020:Context}; SAINT, \cite{Choi2020:Towards}; SAINT+, \cite{Shin2021:Saint+}). Graph-based modeling approaches infer \revinsert{the likelihood with which a certain problem is answered correctly}\revremove{knowledge states} based on the structure induced by question-KC relations (GKT, \cite{Nakagawa2019:Graph}; HGKT, \cite{Tong2020:HGKT}; GIKT, \cite{Yang2020:GIKT}). \revinsert{Unlike BKT techniques and certain logistic regression based approaches such as IRT, deep learning based models often do not provide interpretable parameters or predictions which allow users to quantify a learner's knowledge related to a particular KC. To mitigate this shortcoming recent works have proposed more interpretable network architectures~(\citeNP{Yeung2019:Deep}, \citeNP{Tsutsumi2021:Deep}) and techniques to derive knowledge estimates directly from performance predictions~\cite{Scruggs2020:Extending}.} For an in detail survey on the recent student performance modeling approaches we refer to~\citeNP{Liu2021:Survey}.

Most student performance modeling approaches focus exclusively on question-answering behaviour. A student's sequence of past interactions with the tutoring system is modelled as $\vec{x}_{1:t} = (x_1, \dots, x_t)$. The $t^{\text{th}}$ response is represented by a tuple $x_t = (q_t, a_t)$, where $q_{t}$ is the question (item) identifier and $a_t \in \{0, 1\}$ is binary response correctness. While this formalism has yielded many effective performance modeling techniques, it is limiting in that it assumes the student log data contains only a single type of user interaction: answering questions posed by the ITS. Many aspects of student behaviour such as interactions with learning materials (videos, instructional text), hint usage and information about the current learning context can provide a useful signal, but fall outside the scope of this formalism.

More recently, the EDM community has started exploring alternative types of log data, in an attempt to improve the accuracy of student performance modeling.
\citeNP{Zhang2017:Incorporating} augment DKT with information on response time, attempt number and type of first interaction with the ITS. Later, \citeNP{Yang2018Implicit} enhanced DKT predictions by incorporating more than 10 additional features related to learning context, interaction times and hint usage provided by the \texttt{ASSISTment 2009}~\cite{Feng2009:Addressing} and \texttt{Junyi15}~\cite{Chang2015:Modeling} datasets. While that work employs most information contained in the \texttt{Junyi15} dataset, it fails to utilize the prerequisite structure among topics in the curriculum, and does not evaluate the potential benefit of those features for logistic regression models. EKT~\cite{Liu2019:EKT} uses the question text to learn exercise embeddings which are then used for downstream performance predictions. \revinsert{\citeNP{Eglington2019:Predictiveness} use student response times and correctness rates to cluster the user population into multiple groups each representing a different student phenotype. They then incorporate a set of cluster specific model parameters into a modified PFA model to capture variations between the individual groups leading to improved performance predictions.} MVKM~\cite{Zhao2020:Modeling} uses a multi-view tensor factorization to model knowledge acquisition from different types of learning materials (quizzes, videos, \dots). \revinsert{A recent line of research identified a \emph{Doer} effect which is associated with the finding that interactive problem solving is more indicative for learning outcomes than more passive study activities such as reading and watching lecture videos (\citeNP{Koedinger2015:Learning}, \citeNP{Koedinger2016:Doer}, \citeNP{Koedinger2018:Doer}, \citeNP{Campenhout2021:Doer}).} SAINT+~\cite{Shin2021:Saint+} and MUSE~\cite{Zhang2021:Muse} augment transformer models with interaction time features to capture short-term memorization and forgetting.
Closest to the spirit of this manuscript is an early work by~\citeNP{Feng2009:Addressing}. That work integrates features related to accuracy, response time, attempt-usage and help-seeking behavior into a logistic regression model to enhance exam score predictions. However, their exam score prediction problem is inherently different from our problem of continuous performance modeling because it focuses only on the final learning outcome and does not capture user proficiency on the question and KC level. Unlike our work ~\citeNP{Feng2009:Addressing}, do not evaluate features related to lecture video and reading material consumption, prerequisite structure and learning context. Their study is also limited to two small-scale datasets ($< 1000$ students) collected by the ASSISTment system.

This paper offers the first systematic study of a wide variety of features extracted from alternative types of log data. It analyzes four recent large-scale datasets which capture the learning process at different levels of granularity and focus on various aspects of student behaviour. Our study identifies a set of features which can improve the accuracy of \revinsert{student performance modeling}\revremove{knowledge tracing} techniques and we give recommendations regarding which types of user interactions should be captured in log data of future tutoring systems. Further, our feature evaluation led us to novel logistic regression models achieving a new state-of-the-art performance on all four datasets.

%% file: text/datasets.tex
\section{Datasets}
\label{sec:datasets}

\input{tables/datasets}

In recent years multiple education companies released large-scale ITS datasets to promote research on novel   \revinsert{educational data mining}\revremove{knowledge assessment} techniques. Compared to earlier sizable datasets such as \texttt{Bridge to Algebra 2006}~\cite{Stamper2010:Bridge} and \texttt{ASSISTment 2012}~\cite{Feng2009:Addressing} these new datasets capture an order of magnitude more student responses making them attractive for data intensive modeling approaches.
Table~\ref{tab:datasets} provides an overview of the four recent large-scale student log datasets, from different ITS systems, we use in this paper. Taken together, these datasets capture over 197 million lines of log data from over 789,000 students, including correct/incorrect responses to over 87 million questions. While each ITS collects the conventional timestamp for each question (item) presented to the student, the correctness of their answer, and knowledge component (KC) attributes associated with the question, they also contain additional interaction data. Going beyond pure question-solving activities multiple datasets provide information about how and when students utilize reading materials, lecture videos and hints. In addition there are various types of meta-information about the individual students (e.g. age, gender, \dots) as well as the current learning context  (e.g. school, topic number, \dots). All four tutoring systems exhibit a modular structure in which learning activities are assigned to distinct categories (e.g. pre-test, effective learning, review, \dots). Further, three of the datasets provide meta-information indicating which questions, videos, etc. are related to which KCs.

\begin{figure}[!tbp]
  \small
  \centering
   \begin{minipage}[b]{0.31\textwidth}
     \centering
    \includegraphics[scale=0.34]{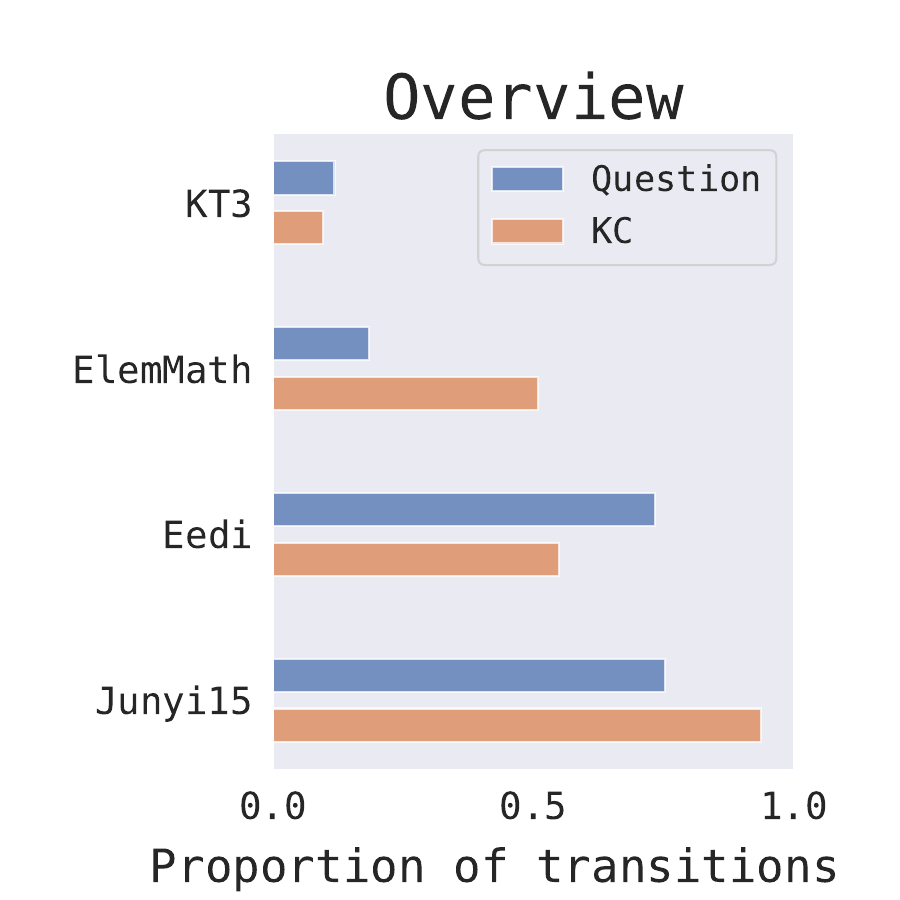}
    \caption{Accuracy when predicting the next question or KC based on the current one. Lower accuracy reflects greater sequence variability across students.}
    \label{fig:entropy}
  \end{minipage}
  \hspace{5mm}
  \begin{minipage}[b]{0.64\textwidth}
    \centering
    \includegraphics[scale=0.34]{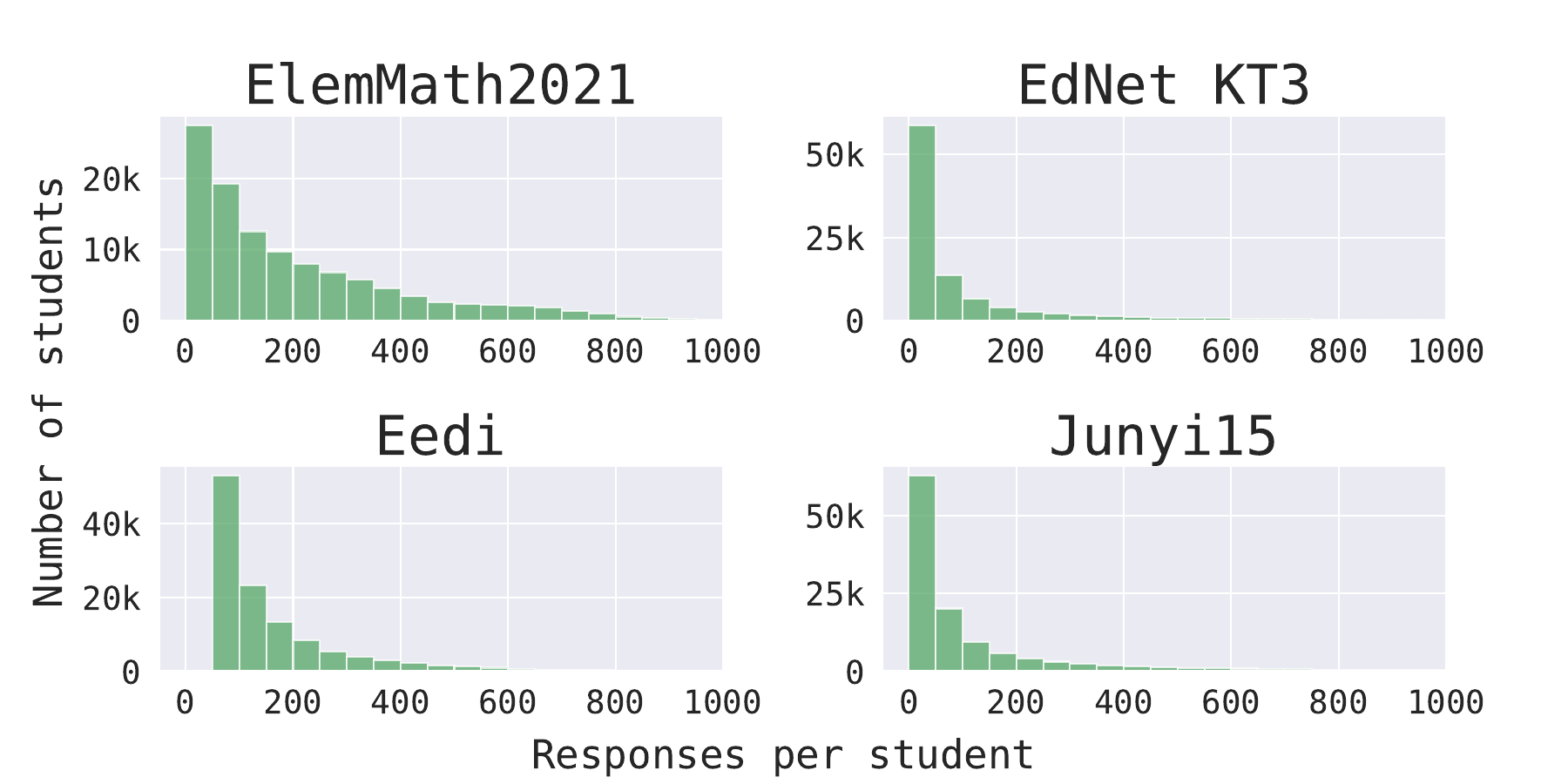}
    \caption{Distribution of the number of responses per student. All four datasets follow a power-law distribution. The \texttt{EdNet KT3} and \texttt{Junyi15} dataset both contain many users with few responses. The users of the \texttt{Eedi} and \texttt{ElemMath2021} tutoring systems have more responses on average. \phantom{xxxxxxxxxxxxxxxxxxxxxxx}} \label{fig:interaction_hist}
  \end{minipage}
\end{figure}

Figure~\ref{fig:interaction_hist} visualizes the distribution of the number of responses per student for each dataset (i.e. the number of answered questions per user). All datasets follow a power-law distribution. The \texttt{EdNet KT3} and \texttt{Junyi15} datasets contain many users with less than 50 completed questions and only a small proportion of users answer more than 100. The \texttt{Eedi} dataset filters out students and questions with less than 50 responses. The \texttt{ElemMath2021} dataset exhibits the largest median response number, and many users answer several hundreds of questions.

Another interesting ITS property is the degree to which all students progress through the study material in a fixed sequence, versus how different and adaptive is the presented sequence across different students. We can get insight into this for each of our datasets by asking how predictable the next question item (or the next KC) is from the previous one, across all students in the dataset. \revinsert{We determine predictability by evaluating the accuracy of a simple model which takes as input the ID of the current question or KC and outputs the ID of the next question or KC which is most likely based on the empirical successor distribution as captured by the log data for the input ID.}
Figure~\ref{fig:entropy} shows how accurately the current question/KC predicts the following question/KC. The \texttt{Junyi15} data \revinsert{exhibits very low variability in its KC sequencing}\revremove{is very static with respect to the KC sequence} and \revinsert{also} tends to present questions in the same order across all students. \texttt{Eedi} logs show more variability in the KC sequence, but the question order is still rather\revremove{static and therefore} predictable. \texttt{ElemMath2021} exhibits moderate KC sequence variations and a highly variable question order. \texttt{EdNet KT3} exhibits the most variation in question and KC order.

Some of the differences between the individual datasets can be traced back to the distinct structures and objectives of the underlying tutoring systems. They teach different subjects, assist students in different ways (K-12 tutoring, standardized test preparation, knowledge diagnosis) and provide students with varying levels of autonomy. We next discuss the individual datasets and corresponding ITS in more detail.

\textbf{EdNet}~\cite{Choi2020:Ednet}: \texttt{EdNet} was introduced as a large-scale benchmarking dataset for \revinsert{student performance modeling}\revremove{knowledge tracing} algorithms. The data was collected over 2 years by Riiid's Santa tutoring system which prepares students in South Korea for the \emph{Test of English for International Communication} (TOEIC\textsuperscript{\textcopyright}) Listening \& Reading. Each test is split into 7 distinct parts, four assessing listening and 3 assessing reading proficiency. Following this structure Santa categorizes its questions into 7 parts (additional fine-grained KCs are provided). Santa is a multi-platform system available on Android, iOS and the web and users have autonomy regarding which test parts they want to focus on. There exist 4 versions of this dataset (\texttt{KT1}, \dots, \texttt{KT4}) capturing student behaviour at increasing levels of detail ranging from pure question answering activity to comprehensive UI interactions. In this paper we analyze the \texttt{EdNet KT3} dataset which contains logs of 297,915 students and provides access to reading and video consumption behaviour. We omit the use of the \texttt{KT4} dataset which augments \texttt{KT3} with purchasing behaviour.

\textbf{Junyi15}~\cite{Chang2015:Modeling}: The Junyi Academy Foundation is an philanthropic organization located in Taiwan. It runs the Junyi Academy online learning platform which offers various educational resources designed for K-12 students. In 2015 the foundation released log data from their mathematics curriculum capturing activities of 247,606 students collected over 2 years. Users of Junyi Academy can choose from a large variety of topics, are able to submit multiple answers to a single question and can request hints. The system registers an answer as correct if no hints are used and the correct answer is submitted on the first attempt. The dataset provides a prerequisite graph which captures semantic dependencies between the individual questions. In addition to a single KC each question is annotated with an area identifier (e.g. algebra, geometry, \dots). In 2020 Junyi Academy shared a newer mathematics dataset on Kaggle~\cite{Pojen2020:Junyi}. Unfortunately, in that dataset each timestamp is rounded to the closest quarter hour which prevents exact reconstruction of the response sequence making it difficult to evaluate student performance modeling algorithms. Because of this we perform our analysis using the \texttt{Junyi15} dataset.

\textbf{Eedi}~\cite{Wang2020:Diagnostic}: Eedi is an UK based online education company which offers a knowledge assessment and misconception diagnosis service. Unlike other ITS which are designed to teach new skills, the Eedi system confronts each student with a series of \textit{diagnostic questions} -- i.e. multiple choice questions in which each incorrect answer is indicative of a common misconception. This process results in a report that helps school teachers to adapt to student specific needs. In 2020 Eedi released a dataset for the NeurIPS Education Challenge~\cite{Wang2021:Results}. It contains mathematics question logs of 118,971 students (primary to high school) and was collected over a 2 year period. Student age and gender as well as information on whether a student qualifies for England's pupil premium grant (a social support program for disadvantaged students) is provided. In contrast to the \texttt{Junyi15} and \texttt{ElemMath2021} datasets which have a prerequisite graph, the \texttt{Eedi} dataset organizes its KCs via a 4-level topic ontology tree. For example the KC \emph{Add and Subtract Vectors} falls under the umbrella of \emph{Basic Vectors} which itself is assigned to \emph{Geometry and Measure} which is connected to the subject \emph{Mathematics}. While analyzing this dataset we noticed that the timestamp information is rounded to the closest minute which prevents exact reconstruction of the interaction sequences. Upon request, the authors provided us with an updated version of the dataset that allows exact recovery of the interaction sequence.

\textbf{Squirrel Ai ElemMath2021}: Squirrel Ai Learning (SQ-Ai) is a K-12 education company located in China which offers individualized after-school tutoring services. SQ-Ai provides their ITS as mobile and Web applications, but also deploys it in over 3000 physical tutoring centers. The centers provide a unique setting in which students can study under the supervision of human teachers and can ask for additional advice and support which augments the ITS's capabilities. Students also have the social experience of working alongside their peers. In this paper we introduce and analyze the \texttt{Squirrel Ai ElemMath2021} dataset. It provides 3 months of behavioral data of 125,246 K-12 students completing various mathematics courses and captures observational data at fine granularity. \texttt{ElemMath2021} gives insight into reading material and lecture video consumption. It also provides meta-information on which learning center and teacher a student is associated with. Each question has a manually assigned difficulty rating ranging from 10 to 90. A prerequisite graph captures dependencies between the individual KCs. Most student learning sessions have a duration of about one hour and the ITS selects a session topic. Each learning process is assigned one of six categories and learning sessions usually follow a pre-test, learning, post-test structure. This makes this dataset particularly interesting because it allows to quantify the learning success of each individual session as the difference between pre-test and post-test performance. 

We conclude this section with a few summarizing remarks. Recent years have yielded large-scale educational datasets which have the potential to fuel future EDM research. The individual datasets exhibit large heterogeneity with regards to the structure and objectives of the underlying tutoring systems as well as captured aspects of the learning process. Motivated by these observations we perform the first comparative evaluation of various student performance modeling techniques across these four large-scale datasets. Further, we use the rich log data to evaluate potential benefits of a variety of alternative features to provide recommendations on which types of observational data is informative for future performance modeling techniques. Our feature evaluation leads us to multiple novel logistic regression models achieving state-of-the-art performance. Given the size of available training data and the structure of the learning processes we also address the question if it is beneficial to train a set of different assessment modules which are specialized on different parts of the ITS.

%% file: tables/datasets.tex
\begin{table}[!ht]
\footnotesize
\centering
\caption{Summary of datasets. Here {\em KC} refers to a knowledge component in the curriculum of the respective system, {\em average correctness} refers to the fraction of questions that were correctly answered across all students, {\em question ID} refers to whether each distinct question (item) had an ID, and {\em KC ID} indicates that each item is linked to at least one KC, {\em platform} indicates if the system logs how students access materials (e.g. mobile app or web browser), {\em social support} indicates if there is information about a student's socioeconomic status, {\em question bundle} indicates if the system groups multiple items into sets which are asked together, {\em elapsed/lag time} indicates whether the data includes detailed timing information allowing the calculation of how long the student took to respond to each question, and the time between presentations of successive questions.  {\em Question difficulty} indicates whether the dataset includes ITS-defined difficulties for each question.  {\em Videos, reading, and hints} indicate whether the dataset provides information about which videos and explanations the student watched or read, and which hints were delivered to them as they worked on practice questions.}
\begin{tabular}{lcccc}
\hline
& \texttt{ElemMath2021} & \texttt{EdNet KT3} & \texttt{Eedi} & \texttt{Junyi15}\\
\hline
\# of students & 125,246 & 297,915 & 118,971 & 247,606 \\
\# of unique questions & 59,892 & 13,169 & 27,613 & 835 \\
\# of KCs & 4,191 & 293 & 388 & 41 \\
\# of logged actions & 62,570,009 & 89,270,654 & 19,834,813 & 25,925,992  \\
\# of student responses & 23,447,961 & 17,954,718 & 19,834,813 & 25,925,992  \\
average correctness & 68.52\% & 66.19\% & 64.30\% & 82.99\% \\
subject & Mathematics & English & Mathematics & Mathematics \\
\hline
timestamp & \yes & \yes & \yes & \yes \\
question ID & \yes & \yes & \yes & \yes \\
KC ID & \yes & \yes & \yes & \yes \\
\hline
age/gender & \noo & \noo & \yes & \noo \\
social support & \noo & \noo & \yes & \noo \\
platform & \yes & \yes & \noo & \noo \\
\hline
teacher/school & \yes & \noo & \yes & \noo \\
study module & \yes & \yes & \yes & \yes \\
pre-requisite graph & \yes & \noo & \noo & \yes \\
KC hierarchy & \noo & \noo & \yes & \noo \\
question bundle & \noo & \yes & \yes & \noo \\
elapsed/lag time & \yes & \yes & \noo & \yes \\
question difficulty & \yes & \noo & \noo & \noo \\
videos  & \yes & \yes & \noo & \noo \\
reading & \yes & \yes & \noo & \noo \\
hints & \noo & \noo & \noo & \yes \\
\hline
\end{tabular}
\label{tab:datasets}
\end{table}

%% file: text/approach.tex
\section{Approach}
\label{sec:approach}

In this paper we examine alternative student performance modeling algorithms using alternative sets of features, across four recent large-scale ITS datasets. Our goals are (1) to discover the features of rich student log data that lead to the most accurate performance models of students, (2) to discover the degree to which the most useful features are task-independent versus dependent on the particular tutoring system, and (3) to discover which types of machine learning algorithms produce the most accurate student models using this log data. We start this Section with a formal definition of the student performance modeling problem. We then discuss prior work on logistic regression based modeling approaches and analyze which types of features they employ. From there, we introduce a set of alternative features leveraging alternative types of log data to offer a foundation for novel \revinsert{student performance prediction}\revremove{knowledge assessment} algorithms. We conclude this Section with the proposal of two additional features which capture long-term and short-term student performance and only rely on response correctness. Appendix~\ref{app:features} provides precise definitions of all features used in our experiments, including implementations details and ITS specific considerations.

\subsection{The Student Performance Modeling Problem}

Student performance modeling is a supervised sequence learning task which traces a student's \revinsert{likelihood to solve different problems}\revremove{latent knowledge state} over time. Reliable \revinsert{performance prediction}\revremove{knowledge assessments} are crucial to enabling ITSs to provide effective individualized feedback to users. More formally, we denote a student's sequence of past interactions with the tutoring system as $\vec{x}_{1:t} = (x_1, \dots, x_t)$. The $t^{\text{th}}$ interaction with the system is represented by the tuple $x_t = (I_t, c_t)$, where $I_t$ indicates the interaction type and $c_t$ is a dataset dependent aggregation of information related to the interaction. In this paper we consider interaction types connected to question answering, video and reading material consumption as well as hint usage. Examples of attributes contained in $c_t$ are timestamp, learning material identifiers, information about the current learning context and student specific features.
Question answering is the most basic interaction type which is monitored by all ITSs. If the $t^{\text{th}}$ interaction is a question response, $c_t$ provides the question (item) identifier $q_{t + 1}$ and binary response correctness $a_t \in \{0, 1\}$. Given a user's history of past interaction with the ITS, the student performance problem is to predict $p(a_{t + 1} = 1 \,|\, q_{t + 1}, \vec{x}_{1:t}$) -- the probability that the student's response will be correct if they are next asked question $q_{t+1}$, given their history $\vec{x}_{1:t}$. In addition to interaction logs, all four datasets provide a knowledge component model which associates each question $q_t$ with a set $KC(q_t)$ containing one or more knowledge components (KCs). Each KC represent a concrete skill which can be\revremove{a} targeted by questions and other learning materials. User interactions are discrete and observed at irregular time intervals. To capture short-term memorization and forgetting it is necessary to utilize additional temporal features. We denote the dependence of variables on the individual student with the subscript $s$.

\subsection{Logistic Regression for Student Performance Modeling}
\label{subsec:logistic}

Logistic regression models enjoy great popularity in the EDM community. At their core each trained regression model takes as input a real-valued feature vector $(\phi_1, \dots, \phi_d)$ that describes the student $s$ and their log data up to this point in the course, along with a question ID.  Note different logistic regression approaches can summarize the student and their history in terms of different features. Each approach calculates its features using its own feature calculation function, $\Phi$ (i.e., $(\phi_1, \dots, \phi_d) = \Phi(q_{s, t+1}\vec{x}_{s,1:t})$.  The trained logistic regression model then uses this feature vector as input, and outputs the probability that the student will correctly answer question $q_{s, t+1}$ if they are asked it at this point in time. The full logistic regression model is therefore of the form
\begin{equation}
  p(a_{s, t+1} = 1 \,|\, q_{s, t+1}, \vec{x}_{s, 1:t}) = \sigma \left( \vec{w}\trans \Phi(q_{s, t+1}, \vec{x}_{s,1:t}) \right).
\end{equation}
Here $\vec{w} \in \mathbb{R}^d$ represents the vector of learned regression weights and $\sigma(x) = 1 / (1 + e^{-x})$ is the sigmoid function which outputs a value between 0 and 1, which is interpreted as the probability that student $s$ will answer question $q_{t+1}$ correctly. A suitable set of weights can be determined by maximizing the likelihood function on the training set. The corresponding maximization problem is usually solved using gradient based optimization algorithms.

Over the years various logistic regression models have been proposed, each employing a distinct feature mapping $\Phi$ to extract a suitable feature set. In this paper we consider Item Response Theory (IRT;~\citeNP{Van2013:Handbook}), in particular a one-parameter version of IRT known as Rasch model~\cite{Rasch1993:Probabilistic}, Performance Factor Analysis (PFA;~\citeNP{Pavlik2009:Performance}), \revinsert{Recent-Performance Factors Analysis (R-PFA;~\citeNP{Galyardt2015:Move}), Predictive Performance Equation (PPE;~\citeNP{Walsh2018:Mechanisms})}, DAS3H~\cite{Choffin2019:DAS3H} and Best-LR~\cite{Gervet2020:Deep}. We now discuss the individual models focusing on how the available interaction data is utilized for their predictions.
First, IRT is the simplest regression model\revremove{s}. It employs a parameter $\alpha_s$ which represents the ability of student $s$ as well as a separate difficulty parameter $\delta_{q}$ for each question (item) $q$. The IRT prediction is defined as
\begin{equation}
\label{eq:irt}
p_{\text{IRT}}(a_{s, t+1} = 1 \,|\, q_{s, t+1}, \vec{x}_{s, 1:t}) = \sigma \left( \alpha_s - \delta_{q_{s, t+1}} \right).
\end{equation}
Unlike IRT, PFA extracts features based on a student's history of past interactions. It computes the number of correct ($c_{s, k}$) and incorrect responses ($f_{s, k}$) prior to the current attempt and introduces a difficulty parameter $\beta_k$ for each individual KC $k$. The PFA prediction is defined as
\begin{equation}
\label{eq:pfa}
p_{\text{PFA}}(a_{s, t+1} = 1 \,|\, q_{s, t+1}, \vec{x}_{s, 1:t}) = \sigma \left( \sum_{k \in KC(q_{s, t+1})} \beta_k + \gamma_k c_{s,k} + \rho_k f_{s,k} \right).
\end{equation}
\revinsert{R-PFA is motivated by the idea that more recently observed student responses are more indicative for future performance than older ones. R-PFA builds on PFA by introducing two features which for each KC $k$ look at all interactions of student $s$ with $k$ up to time $t$ and computes: (i) A recency-weighted count of previous failures $F_{s, k, t}$ using exponential decay. (ii) A recency-weighted proportion of past successes $R_{s, k, t}$ using normalized exponential decay. The degree of decay is controlled by the hyperparameters $d_F$ and $d_R \in [0, 1]$. To allow the computation of $R_{s, k, t}$ when a student visits a KC for the first time, R-PFA appends their interaction history with $k$ with $g = 3$ incorrect ``ghost attempt''. We denote the total number of responses of student $s$ related to KC $k$ as $a_{s,k}$ and use a correctness indicator $a_{{s, k, i}}$ which is 1 when $s$'s $i$-th attempt on KC $k$ was correct and 0 otherwise. The R-PFA prediction is defined as}
\begin{align}
\begin{split}
p_{\text{R-PFA}}(a_{s, t+1} = 1 \,|\, q_{s, t+1}, \vec{x}_{s, 1:t}) &= \sigma \left( \sum_{k \in KC(q_{s, t+1})} \beta_k + \gamma_k F_{s, k, t} + \rho_k R_{s, k, t} \right) \\
F_{s, k, t} = \sum_{i = 1}^{a_{s,k}} d_F^{(a_{s,k} + 1) - i} (1 - a_{{s, k, i}})&, \quad
R_{s, k, t} = \sum_{i = (1 - g)}^{a_{s,k}} \frac{d_R^{a_{s,k} - i}}{\sum_{j = (1 - g)}^{a_{s,k}} d_R^{a_{s,k} - i}} a_{{s, k, i}}\,.
\end{split}
\end{align}
\revinsert{In the context of word pair learning, PPE was proposed to capture the spacing effect~\cite{Cepeda2008:Spacing} -- the phenomena that spaced out practice repetitions slow down learning but increase retention rates -- by introducing a weighting scheme that considers the delay between individual practice sessions. PPE assumes a multiplicative relationship between the number of prior attempts $a_{s,k}$ with a time variable $T_k$. The model features a learning rate parameter $c$ and the forgetting rate is controlled by parameters $x$, $b$ and $m$. These four hyperparameters need to be set by the user. We define $\Delta_{s,k,i}$ to be the real time passed since student $s$'s $i$-th response to KC $k$. The PPE prediction is defined as}
\begin{align}
\label{eq:ppe}
\begin{split}
p_{\text{PPE}}(a_{s, t+1} = 1 \,|\, q_{s, t+1}, \vec{x}_{s, 1:t}) &= \sigma \left( \sum_{k \in KC(q_{s, t+1})} \beta_k + \gamma_k \left(a_{s,k}^c T_k^{-d_t} \right) \right) \\
T_k = \left( \sum_{i = 1}^{a_{s,k}} \Delta_{s,k,i}^{1 - x} \right) \left( \sum_{i = 1}^{a_{s,k}} \frac{1}{\Delta_{s,k,j}^{-x}} \right)&, \quad
d_t = b + m \left( \frac{1}{a_{s,k}} \sum_{i = 1}^{a_{s,k}} \frac{1}{\ln(\Delta_{s,k,i} - \Delta_{s,k,i + 1} + e)} \right).
\end{split}
\end{align}
DAS3H is a more recent model which combines aspects of IRT and PFA and extends them with time-window based count features. It defines a set of time windows $W = \{ 1/24, 1, 7, 30, +\infty \}$ measured in days. For each window $w \in W$, DAS3H determines the number of prior correct responses ($c_{s, k, w}$) and overall attempts ($a_{s, k, w}$) of student $s$ on KC $k$ which fall into the window. A scaling function $\phi(x) = \log(1 + x)$ is applied to avoid features of large magnitude. The DAS3H prediction is defined as
\begin{align}
\begin{split}
  p_{\text{DAS3H}}(a_{s, t+1} = 1 \,|\, q_{s, t+1}, \vec{x}_{s, 1:t}) = &\sigma \Bigg( \alpha_s - \delta_{q_{s, t+1}} + \sum_{k \in KC(q_{s, t+1})} \beta_k + \\
  &\sum_{k \in KC(q_{s, t+1})} \sum_{w = 0}^{W - 1} \theta_{k, 2w + 1} \phi(c_{s,k,w}) - \theta_{k, 2w + 2} \phi(a_{s,k,w}) \Bigg).
\end{split}
\end{align}
\citeNP{Gervet2020:Deep} performed a comparative evaluation of student performance modeling algorithms across 9 real-world datasets. They also evaluated the effects of question, KC and total count as well as time window based count features leading them to a new logistic regression model referred to as Best-LR. Best-LR is similar to DAS3H, but does not use time-window features and uses $c_s$ and $f_s$ as additional features that capture the total number of prior correct and incorrect responses. The Best-LR prediction is defined as
\begin{align}
\label{eq:bestlr}
\begin{split}
  p_{\text{Best-LR}}(a_{s, t+1} = 1 \,|\, q_{s, t+1}, \vec{x}_{s, 1:t}) = &\sigma \Bigg( \alpha_s - \delta_{q_{s, t+1}} + \phi(c_s) + \phi(f_s) + \\
  &\sum_{k \in KC(q_{s, t+1})} \beta_k + \gamma_k \phi(c_{s, k}) + \rho_k \phi(f_{s, k}) \Bigg).
\end{split}
\end{align}

Overall, we can see that these four logistic regression models are mainly based on two types of features: (i) One-hot encodings that allow the models to infer question and KC specific difficulty; (ii) Count based features that summarize a student's past interaction history with the system computed at various level of granularity (total/KC/question-level) potentially augmented with time windows \revinsert{and time-based weighting schemes} to introduce a temporal dimension in the predictions.
Looking back at the dataset discussion provided in Section~\ref{sec:datasets}, we note that the current feature sets only use a small fraction of the information collected by the tutoring systems. In the following we aim at increasing the amount of utilized information by exploring alternative types of features which can be extracted from the log data and can serve as foundation for future student performance modeling algorithms.

\subsection{Features Based on Rich Observational Data}
\label{subsec:rich}

Tutoring systems collect a wide variety of observational data during the learning process. Here we discuss a range of alternative features leveraging alternative types of log data. The individual features can be used to augment the logistic regression models discussed in Subsection~\ref{subsec:logistic}, but might also be combined with deep learning based modeling techniques. As shown by Table~\ref{tab:datasets}, each dataset captures different aspects of the learning process and supports a different subset of the discussed features.

\textbf{Temporal features}: Many performance modeling techniques treat student interactions as discrete tokens and omit timestamp information which can be indicators of cognitive processes such as short-term memorization and forgetting. DAS3H uses time window based count features to summarize the user history. Here we discuss two additional types of temporal features: (i) One-hot encoded datetime and (ii) the interaction time based features introduced by~\citeNP{Shin2021:Saint+}.
By providing the model with information about the specific week and month of interaction we try to capture effects of school work outside the ITS on \revinsert{student learning}\revremove{knowledge development}. The hour and day encodings aim at tracing temporal effects on a smaller timescale. For example, students might produce more incorrect responses when studying late at night.
Recently, \citeNP{Shin2021:Saint+} introduced a deep learning approach which employs \emph{elapsed time} and \emph{lag time} features to capture temporal aspects of student behaviour during question solving activities. Elapsed time measures the time span from question display to response submission. The idea is that a faster response is correlated with student proficiency. Lag time measures the time passed between the completion of the previous exercise until the next question is received. Lag time can be indicative for short-term memorization and forgetting.
\revinsert{The lag time between questions can be affected by student behaviour and system design choices. Some examples are displayed explanations after incorrect responses, pop-up messages and study breaks.}
For our experiments we convert elapsed time and lag time values, $x$, to scaled values $\phi(x) = \log(1 + x)$ and also use the categorical one-hot encodings from \citeNP{Shin2021:Saint+}. There, elapsed time is capped off at 300 seconds and categorized based on the integer second. Lag time is rounded to integer minutes and assigned to one of 150 categories ($0, 1, 2, 3 ,4, 5, 10, 20, 30, \dots, 1440$). We evaluate two variations of the features. In the first version we compute elapsed time and lag time values based on interactions with the \emph{current} question. In the second version we compute them based on interactions with the \emph{prior} question. Because it is unknown how long a student will take to answer a question before the question is asked, we cannot realistically use this elapsed time feature for predicting correctness in answering a new question. Therefore, we omit the elapsed time feature for the current question in our experiments described in Subsection~\ref{subsec:integrating_features}.  

\textbf{Learning context}: Knowledge about the context a learning activity is placed in can be informative for performance predictions. For example, all four datasets considered in this work group learning activities into distinct context categories we call study modules (e.g. pre-test, effective learning, review, \dots).  Here effective learning study modules try to teach novel KCs, whereas review study modules aim at deepening proficiency related to KCs the student is already familiar with. Providing a student performance model with information about the corresponding study module can help it adapt its predictions to these different contexts. Additional context information is provided by the exercise structure. For example, the \texttt{ElemMath2021} dataset marks each interaction with a course and topic identifier and offers a manually curated question difficulty score. The \texttt{EdNet} dataset labels questions based on which part of the TOIEC exam they address and groups them into bundles--a bundle is a set of questions which is asked together. Large parts of the \texttt{ElemMath2021} dataset were collected in physical learning centers where students can study under the supervision of human teachers and each interaction contains school and teacher identifiers. This information can be useful because students visiting the same learning center might exhibit similar strengths and weaknesses. In Section~\ref{sec:experiments} we evaluate the potential benefits of learning context information for student performance modeling by encoding the categorical information into one-hot vectors and passing it into logistic regression models. Additionally, we evaluate count features defined for the individual study modules and \texttt{EdNet} parts.

\textbf{Personal context}: By collecting a student's personal attributes a tutoring system can offer a curriculum adapted to their individual needs. For example, information about a student's age and grade can be used to select suitable learning materials. Further, knowledge about personal attributes can enable an ITS to serve as a vehicle for research on educational outcomes. For example, it is a well studied phenomena that socioeconomic background correlates with educational attainment (see for example~\citeNP{White1982:Relation}, \citeNP{Aikens2008:Socioeconomic}, \citeNP{Stumm2017:Socioeconomic}). A related research question is how ITSs can be used to narrow the achievement gap~\cite{Huang2016:Intelligent}. While both \texttt{ElemMath2021} and \texttt{EdNet} datasets indicate which modality students use to access the ITS, the \texttt{Eedi} dataset is the only one that provides more detailed personal information in form of age, gender and socioeconomic status. Features extracted from these sensitive attributes need to be handled with care to avoid discrimination against individual groups of students. Outside of student performance modeling these features might be used to detect existing biases in the system. We evaluate one-hot encodings related to age, gender and socioeconomic status.

\textbf{KC graph features}: In addition to a KC model three of the datasets provide a graph structure capturing semantic dependencies between the individual KCs and questions. We can leverage these graph structures by defining \emph{prerequisite} and \emph{postrequisite} count features. These features compute a student's number of prior attempts and correct responses on KCs which are prerequisite and postrequisite to the KCs of the current question. The motivation for these features is that the mastery of related KCs can carry over to the current question. For example, it is very likely that a user that is proficient in multi-digit multiplication can solve single-digit multiplication exercises as well. The \texttt{Eedi} dataset provides a 4-level KC ontology tree which associates each question with a leaf. Here we compute a prerequisite feature by counting the number of interactions related to the direct parent node. We also investigate the use of sparse vectors taken from the KC graph adjacency matrix as an alternative way to incorporate the semantic information into our models.

\textbf{Learning material consumption}: Modern tutoring system offer a rich variety of learning materials in form of video lectures, audio recordings and written explanations. Information on how students interact with the available resources can benefit performance predictions. For example, watching a video lecture introducing a new KC can foster a student's understanding before they enter the next exercise session. The \texttt{ElemMath2021} and \texttt{Ednet KT3} datasets record interactions with lecture video and written explanations. Count features capture the number of videos and texts a user has interacted with prior to the current question and can be computed on a per KC and an overall level. Video and reading time in minutes might also provide a useful signal. Another feature which can be indicative for proficiency is the number of videos a student skips. The \texttt{Junyi15} datasets provides us with information on hint usage. We experiment with features capturing the number of used hints as well as minutes spent on reading hints aggregated on a per KC and an overall level.

\subsection{Features Based on Response Correctness}
\label{subsec:response}

\subsubsection{Smoothed Average Correctness}
\label{subsec:smoothed_average_correctness}

The performance of logistic regression models relies on a feature set which offers a suitable representation of the problem domain. Unlike neural network based approaches, linear and linear logistic regression models are unable to recover higher-order dependencies between individual features on their own. For example, given two separate features describing a student's number of prior correct responses, $c$, and overall attempts, $a$, the linear logistic model is unable to infer the ratio $r = c/a$ of average correctness which can be a valuable indicator of a student's long-term performance. To mitigate this particular issue we introduce an additional feature, $r_s$ capturing average correctness of student $s$ over time as
\begin{equation}
  r_{s} = \frac{c_s + \eta \bar{r}}{a_s + \eta}
\end{equation}
where $a_s$ is the number of questions attempted by student $s$, and $c_s$ is the number of questions the student answered correctly.   Here, $\bar{r}$ is the average correctness rate over all other students in the dataset, and $\eta \in \mathbb{N}$ is a smoothing parameter which biases the estimated average correctness rate, $\bar{r}_s$ of student $s$ towards this all students average $\bar{r}$. The use of smoothing reduces the feature variance during a student's  initial interactions with the ITS.
\revinsert{Related to our approach R-PFA~\cite{Galyardt2015:Move} introduced a feature that computes a recency-weighted proportion of correct responses for each individual KC via an exponential decay function. The LKT framework~\citeNP{Pavlik2020:Logistic} provides multiple features that can trace the ratio of correct student responses. By implementing multiple recency-based decay functions LKT extends R-PFA and introduces features that capture the overall correctness rate of each individual student. Unlike our formulation LKT does not allow the use of a smoothing parameter to reduce the feature variance by biasing towards the all student average.}\revremove{A prior work by~\citeNP{Pavlik2020:Logistic} proposed, but did not evaluate, an average correctness feature without the smoothing parameter for student performance modeling.}
\revinsert{We calibrated the smoothing parameter for our experiments by evaluation $\eta \in \{0, 1, 5, 10, 25, 50, 100, 250\}$ using the \texttt{ElemMath2021} dataset. Smoothing yielded benefits and we settled on $\eta = 5$ for which we observed the largest accuracy improvements.}
\revremove{While calibrating this parameter for our experiments, we observed benefits from smoothing and settled on $\eta = 5$.}

\subsubsection{Response Patterns}
\label{subsec:response_pattern}
\begin{figure}[!tbp]
  \small
  \centering
  \includegraphics[width=\textwidth]{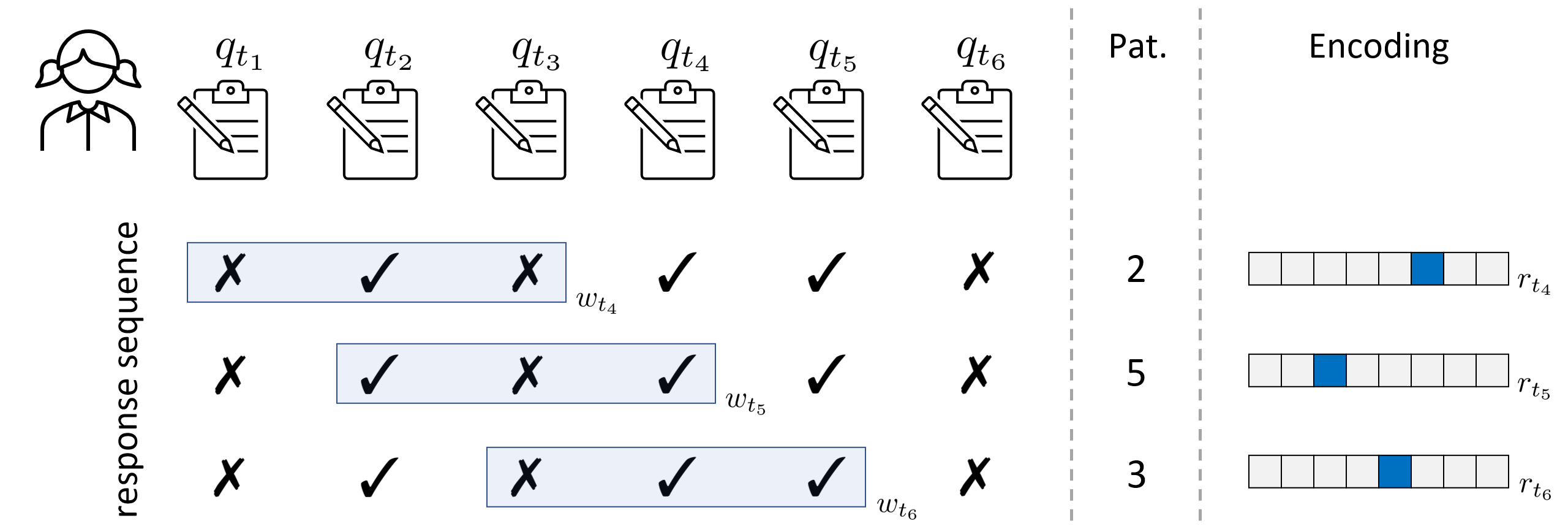}
   \caption{A response pattern $r_t$ is a one-hot encoding vector which represents the binary sequence $w_t$ formed by a student's $n$ most recent correct and incorrect responses. Logistic regression models can use response patterns to infer aspects of a student's short-term behaviour.}
  \label{fig:response_pattern}
\end{figure}

Student performance modeling approaches that employ recurrent or transformer based neural networks take interaction sequences describing \emph{multiple} student responses as input (e.g. the most recent 100). Provided this time-series data, deep learning algorithms are able to discover patterns and temporal dependencies in student behaviour without requiring additional human feature engineering. Among the logistic regression models discussed in Subsection~\ref{subsec:logistic} only DAS3H incorporates temporal information in form of time window based count features to summarize student interactions over time scales from one hour to one months. While DAS3H can capture effects of cognitive processes such as long-term memorization and forgetting, it is still at disadvantage compared to deep learning based approaches that can also infer aspects of student behaviour occurring on smaller time-scales. Indeed, it has been shown that a student's most recent responses have a large effect on DKT performance predictions~(\citeNP{Ding2019:Deep}, \citeNP{Ding2021:Interpretability}).

Here, inspired by the use of n-gram models in the NLP community (e.g.~\citeNP{Manning1999:Foundations}), we propose \emph{response patterns} as a feature which allows logistic regression models to infer aspects impacting short-term student performance. At time $t$, a response pattern $r_t \in \mathbb{R}^{2^n}$ is defined as a one-hot encoded vector that represents a student's sequence of $n \in \mathbb{N}$ most recent responses $w_t = (a_{t - n}, \dots, a_{t - 1})$ formed by binary correctness indicators $a_{t - n}, \dots, a_{t - 1} \in \{0, 1\}$. The encoding process is visualized by Figure~\ref{fig:response_pattern}. \revinsert{For our experiments we calibrated the pattern length by evaluating $n \in \{1, 2, 3, \dots, 14\}$ using the \texttt{ElemMath2021} dataset. We settled on $n = 10$ for which we observed the largest accuracy improvements.}
Response patterns are designed to allow logistic regression models to capture momentum in student performance. They allow the model to infer how challenging the current exercise session is for the user and can also be indicative for question skipping behaviour. In Subsection~\ref{subsec:integrating_features} we combine response patterns, smoothed average correctness and DAS3H time window features to propose Best-LR+, a regression model which offers performance competitive to deep learning based techniques while only relying on information related to response correctness.

%% file: text/experiments.tex
\section{Experiments}
\label{sec:experiments}

We evaluate the benefit of alternative features extracted from different types of log data for student performance modeling using the four large-scale datasets discussed in Section~\ref{sec:datasets}. After an initial feature evaluation, we combine helpful features to form novel state-of-the-art logistic regression models.
Motivated by the size of the recent datasets, we also investigate two ways of partitioning the individual datasets to train multiple assessment models targeting different parts of the learning process. First, we show how \revinsert{partitioning }\revremove{splitting the}\revinsert{a student's} interaction sequence\revremove{s} by response number can be used to train \emph{time-specialized} models \revinsert{-- focusing on responses submitted earlier or later in the interaction sequence -- to}\revremove{which can} mitigate the cold start problem~(\citeNP{Gervet2020:Deep}, \citeNP{Zhang2021:Knowledge}). We then analyze how features describing the learning context (e.g. study module, topic, course, \dots) can be used to train multiple \emph{context-specialized} models whose individual predictions can be combined to improve overall prediction quality even further.

\subsection{Evaluation Methodology}

\begin{figure}[!t]
  \small
  \centering
   \begin{minipage}[b]{0.475\textwidth}
    \centering
    \includegraphics[scale=0.45]{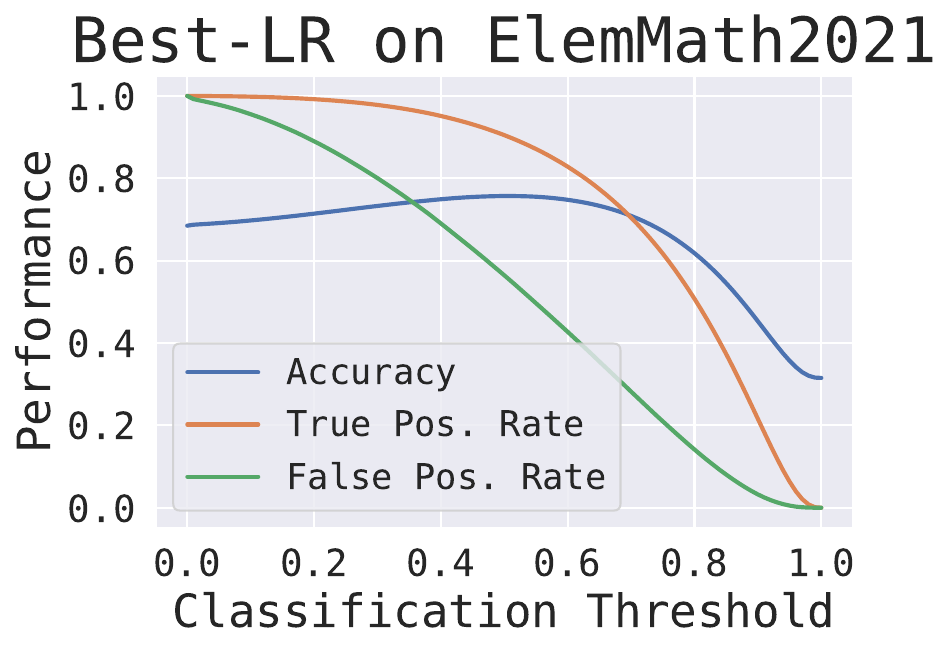}
    \caption{Accuracy, true positive rate and false positive rate of a Best-LR model trained on the \texttt{ElemMath2021} dataset for different classification thresholds. The classification threshold converts the probability output by the model into a binary correct/incorrect prediction.}
    \label{fig:tpr_fpr}
  \end{minipage}
  \hfill
  \begin{minipage}[b]{0.475\textwidth}
    \centering
    \includegraphics[scale=0.45]{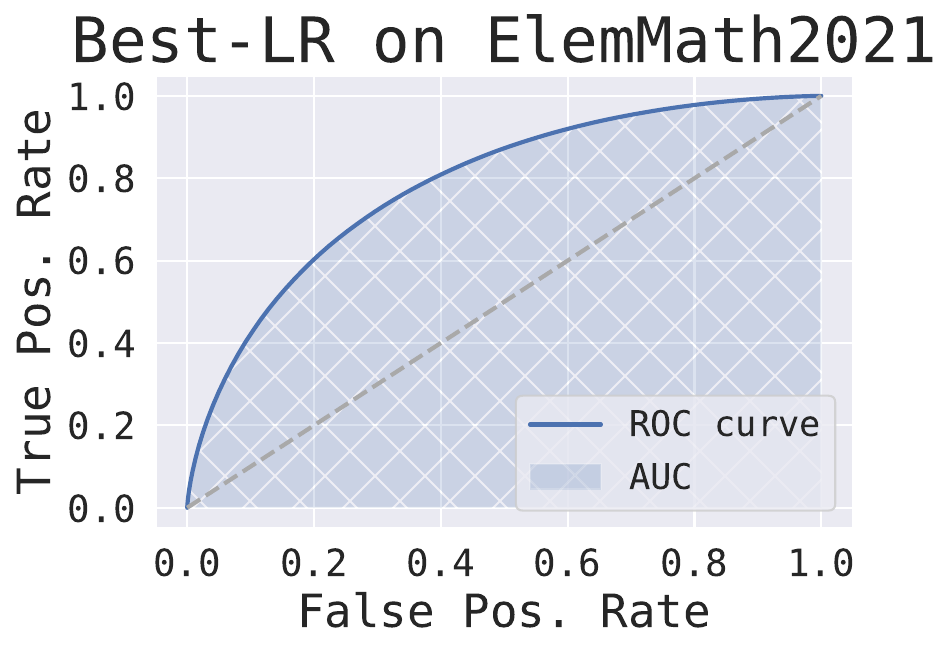}
    \caption{The receiver operating characteristic (ROC) curve captures the relationship between true and false positive rates as the classification threshold is swept from 0 to 1. The AUC score stands for the Area Under the Curve. A perfect system will have an AUC of 1.0.} 
    \label{fig:roc}
  \end{minipage}
\end{figure}

To be in line with prior work, we start data preparation by filtering out students with less than ten answered questions~(\citeNP{Pandey2019:Self}, \citeNP{Choffin2019:DAS3H}, \citeNP{Piech2015:Deep}, \citeNP{Gervet2020:Deep}). The \texttt{Ednet KT3} and \texttt{Eedi} dataset both contain questions annotated with multiple KCs which yields difficulties for some of the deep learning baselines (DKT and SAKT). In those cases we introduce new artificial KCs to represent each unique combination of original KCs. In all our experiments we perform a 5-fold cross-validation on the student level. In each fold 80\% of the students are used for training and parameter selection and 20\% are used for testing. Thus, all test results are obtained only from predictions over student who were not observed during model training. We report performance in terms of prediction accuracy (ACC) and area under curve (AUC).
The AUC score captures the area under the receiver operating characteristic (ROC) curve and is a popular evaluation metric for \revinsert{student performance prediction}\revremove{knowledge tracing} algorithms. The ROC curve plots the true-positive rate against the false-positive rate at all decision thresholds. One way of interpreting the AUC score is viewing it as the probability of assigning a random correct student response a higher probability of correctness than a random incorrect response. As concrete examples we visualize the predictive performance and ROC curve for a Best-LR model trained on the \texttt{ElemMath2021} dataset in Figures~\ref{fig:tpr_fpr} and~\ref{fig:roc}.

For the computation of average correctness and response pattern features we set smoothing parameter $\eta = 5$ and sequence length $n = 10$ respectively \revinsert{(discussed in Sections~\ref{subsec:smoothed_average_correctness} and~\ref{subsec:response_pattern})}. \revinsert{For R-PFA we followed~\cite{Galyardt2015:Move} by fixing the number of ghost attempts $g = 3$ and determined decay rate parameters $d_F$ and $d_R$ by selecting the best value in $\{0.1, \dots, 0.9, 1.0\}$ for each dataset. For PPE we followed~\cite{Walsh2018:Mechanisms} by fixing $c = 0.1$ and $x = 0.6$ and evaluated 20 equally-spaced decay parameters $b \in [0.01, 0.05]$ and $m \in [0.02, 0.04]$ for each dataset.} Additional implementation details of the different features as well as ITS specific considerations are provided in Appendix~\ref{app:features}.
For the evaluation of the deep learning models, we \revinsert{performed a grid-search and defined hyperparameter search spaces which extend}\revremove{relied on} the hyperparameters \revinsert{selected}\revremove{reported} in the cited references~(DKT, \cite{Piech2015:Deep}; SAKT, \cite{Pandey2019:Self}; SAINT, \cite{Choi2020:Towards}; SAINT+, \cite{Shin2021:Saint+}). A detailed list of the used\revremove{model} hyperparameter\revremove{s} \revinsert{search spaces} is provided in Appendix~\ref{app:hyperparam}. All models were trained for 100 epochs without learning rate decay.
For our logistic regression experiments we rely on Scikit-learn~\cite{Pedregosa2011:Scikit} and all deep learning architectures were implemented using PyTorch~\cite{Paszke2019:Pytorch}. Our implementation of the regression models uses combinations of attempt and correct count features which is a slight deviation from the original PFA and Best-LR formulations which count the number of prior correct and incorrect responses. While the individual features have a different interpretation, the two feature pairs are collinear to each other and provide identical information to the model. The code to reproduce our experiments as well as links to\revremove{all} used datasets are shared on GitHub\footnote{\url{https://github.com/rschmucker/Large-Scale-Knowledge-Tracing}}.

\subsection{Utility of Individual Features with Logistic Regression}
\label{subsec:feature_significance}

To evaluate the effectiveness of the different features discussed in Section~\ref{sec:approach} we perform two experiments in the context of logistic regression modeling. In the first experiment we analyze each feature by training a logistic regression model using only that feature and a bias (constant) term. This allows us to quantify the predictive power of each feature in isolation. In the second experiment we evaluate the marginal utility gained from each feature by training an augmented Best-LR model (Eq.~\ref{eq:bestlr}) based on the Best-LR feature set plus the additional feature. This is particularly interesting because the Best-LR feature set was created by combining features proposed by various prior regression models and it achieves state-of-the-art performance on multiple educational datasets~\cite{Gervet2020:Deep}. If a feature can provide marginal utility on top of the Best-LR feature set it is likely that it can contribute to other \revinsert{student performance modeling}\revremove{knowledge tracing} techniques as well and should be captured by the ITS.

\input{tables/individual_feature_evaluation}

ACC and AUC scores achieved by the logistic regression models trained on the \textit{individual features} in isolation are shown in Table~\ref{tab:single_feature_performance}. Somewhat unsurprisingly, the question (item) identifier -- representing the main feature used by IRT (EQ~\ref{eq:irt}) -- is effective for all datasets and is the single most predictive feature for \texttt{ElemMath2021} and \texttt{EdNet KT3}. KC identifiers are very informative as well, but to a lesser extent than the question identifiers. This indicates varying difficulty among questions targeting the same KC. The PFA model (EQ~\ref{eq:pfa}) which only estimates KC difficulties is unable to capture these variations on the question-level. The attempt and correct count features are informative for all datasets and there is benefit in using total-, KC- and question-level counts in combination. Question-level count features are informative for \texttt{Junyi15} which is the only dataset that frequently presents users previously encountered questions. \revinsert{Combined R-PFA's two recency-weighted features deliver more accurate performance predictions than PPE's spacing feature.} DAS3H introduced count features computed for different time windows (TWs). TW based count features are an extension of the standard count features and yield additional benefits for all four settings. The elapsed time and lag time features are based on user interaction times and are both informative. The datetime features describing month, week, day and hour of interaction yield little signal on their own. 

\revinsert{Looking at the individual learning context features we observe varying utility.}\revremove{The features describing the learning context vary in utility}. The study module feature is informative for all datasets, as are the counts of student correct responses and attempts associated with different study modules. Predictions for the \texttt{Eedi} dataset benefit largely from information about the specific group a student belongs to\revremove{o}. Reasons for this could be varying group proficiency levels and differences in diagnostic test difficulty. Bundle and part identifiers both provide a useful signal for \texttt{EdNet KT3} and \texttt{Eedi}. Features describing personal information have little predictive power on their own. Among this family of features, social support yields the largest AUC score improvement over the always correct baseline. Features extracted from the prerequisite graph are effective for the three datasets that support them. The logistic regression models trained on prerequisite and postrequisite count features exhibit the best AUC scores on \texttt{Junyi15}. The features describing study material consumption all provide a predictive signal, but do not yield enough information for accurate \revinsert{performance predictions}\revremove{knowledge assessments} on their own. The smoothed average correctness and response pattern features lead to good performance on all four datasets.

\input{tables/best_lr_feature_evaluation}

Moving forward we evaluate the marginal utility of each feature by evaluating the performance of logistic regression models trained using the \textit{Best-LR feature set} augmented with this particular feature. Table~\ref{tab:bestlr_feature_performance} shows the results of this experiment. \revinsert{The combination of the two R-PFA features is beneficial for all four datasets and increases the AUC score for \texttt{Junyi15} by $1.75\%$. The PPE feature is also helpful, but to a lesser extend than the R-PFA features.} Time window based count features lead to improved \revinsert{predictions}\revremove{knowledge assessments} for all datasets. In combination, they improve the AUC scores of \texttt{EdNet KT3} and \texttt{Junyi} by $1.29\%$ and $2.2\%$ respectively. The elapsed time and lag time features offer a useful signal for the three datasets which capture temporal interaction data. The datetime features provide little utility in all cases (the best observed value is a $0.08\%$ AUC improvement for \texttt{Junyi15}). 

\revinsert{Context} information about the study module a question is placed in is beneficial in all settings and leads to the largest AUC score improvement \revinsert{over the Best-LR baseline model} for \texttt{Junyi15}. Indication of the group a student belongs to improves model performance for \texttt{Eedi}. Knowing which topic a student is currently visiting is a useful signal for \texttt{ElemMath2021}. Information related to bundle and part structure improves \texttt{EdNet KT3} and \texttt{Eedi} predictions. \texttt{ElemMath2021}'s manually assigned difficulty ratings lead to no substantial improvements likely due to the fact that the Best-LR feature set allows the model to learn question difficulty on its own. The features describing a student's personal information provide little marginal utility. Count features derived from the question/KC prerequisite graphs yield a sizeable improvement in assessment quality. Features targeting study material consumption yield some marginal utility when available. Discrete learning material consumption counts lead to larger improvements than the continuous time features. The smoothed average correctness feature leads to noticeable improvements for all four datasets. The response pattern feature enhances assessments in all cases and yields the largest improvements for the $\texttt{ElemMath2021}$ and $\texttt{Eedi}$ dataset (over $0.5\%$ AUC). This shows that the last few most recent student responses yield a valuable signal for performance prediction.

Overall, we observe that a variety of alternative features derived from different types of log data can enhance student performance modeling. Tutoring systems that track temporal aspects of student behaviour in detail can employ elapsed time and lag time features. Additional information about the current learning context is valuable and should be captured. While the study module features improve predictions for all datasets, other ITS specific context features such as group, topic and bundle identifiers vary in utility. The count features derived from the provided prerequisite and hierarchical graph structures increased prediction quality in all cases. Log data related to study material consumption also provides a useful signal. Beyond student performance modeling this type of log data might also prove itself useful for evaluating the effects of individual video lectures and written explanations in future work. Lastly, the smoothed average correctness and response pattern features that only require answer correctness improve predictions for all four tutoring systems substantially.

\subsection{Integrating Features into Logistic Regression}
\label{subsec:integrating_features}

We have seen how individual features can be integrated into the Best-LR model to increase prediction quality. We now experiment with augmented logistic regression models that incorporate combinations of \emph{multiple} beneficial features. The number of parameters learned by the different approaches is provided in Table~\ref{tab:num_params}. Because each dataset provides different types of log data it only supports a subset of the explored features. The first model we propose and evaluate is Best-LR+.
\begin{quote}
 {\bf Best-LR+ method}. This method augments the Best-LR method (the current state-of-the-art logistic regression method (EQ~\ref{eq:bestlr})) by adding \revinsert{R-PFA's two recency-weighted features, PPE's spacing time feature and}\revremove{the} DAS3H\revinsert{'s} time window-based count features on total-, KC- and question-level as well as smoothed average correctness and response pattern features. All of these features can be calculated using only the raw question-response information from the student log data. 
 
\end{quote}
Note that because the features used in Best-LR+ rely only on question-response log data, this method can be applied to all four datasets, as well as many earlier datasets that lack new features available in the four we study here. We define Best-LR+ in this way, to explore whether tutoring systems that log only this question-response data can improve the accuracy of the student performance modeling by calculating and adding in these count features.

We further propose dataset specific augmented logistic regression models (AugmentedLR) by combining the helpful features marked in Table~\ref{tab:bestlr_feature_performance}.
\begin{quote}
 {\bf AugmentedLR method}. This logistic regression method uses all of the features employed in the Best-LR model, and adds in all of the marked features from Table~\ref{tab:bestlr_feature_performance}, which were found to individually provide \revinsert{AUC score} improvements \revinsert{of more than $0.05\%$} over the state-of-the-art results of Best-LR.  
\end{quote}
Note that the AugmentedLR method employs a superset of the features used by Best-LR+. Because the features used in AugmentedLR go beyond simple logs of questions and responses, this method can only be applied to datasets that capture this additional information. We define AugmentedLR in this way, to explore whether ITS systems that do not yet log these augmented features can improve the accuracy of their student performance modeling by capturing and utilizing these additional features in their student log data. Note that some of the features marked in Table~\ref{tab:bestlr_feature_performance} and used by AugmentedLR are available in only a subset of our four datasets. \revinsert{During our AugmentedLR feature selection we made two additions to the over $0.05\%$ AUC improvement rule to mitigate redundancy: (i) In cases where count and ID feature target the same attribute we select the one yielding the larger AUC improvements. (ii) Information about the current lag time is preferred over prior lag time.}

Experimental results for these two proposed logistic regression models are presented in Table~\ref{tab:model_comparison}, along with results for previously published logistic regression approaches IRT, PFA, \revinsert{R-PFA, PPE,} DAS3H, Best-LR and previously published deep learning approaches DKT, SAKT, SAINT and SAINT+.

Examining the results, first compare the results for Best-LR+ to the previous state of the art in logistic regression methods, Best-LR. Notice that Best-LR+ benefits from its additional question-response features and outperforms Best-LR in all four datasets. On \texttt{EdNet KT3} and \texttt{Juni15} Best-LR+ improves the AUC scores of the previous best logistic regression models (Best-LR) by more than \revinsert{$1.7\%$}\revremove{$1.5\%$}. These results suggest that the additional features used by Best-LR+ should be incorporated into student performance modeling in any ITS, even those for which the student log data contains only the sequence of question-response pairs.

Examining the results for AugmentedLR, it is apparent that this provides the most accurate method for student modeling across these four diverse datasets, among the logistic regression methods considered here. Even when considering recent deep learning based models, the AugmentedLR models outperform all logistic regression and deep neural network methods on all datasets except \texttt{Eedi} where the deep network method DKT leads to the best performance predictions.  Especially on the \texttt{Junyi15} dataset AugmentedLR increases ACC by \revinsert{$2.1\%$}\revremove{$2\%$} and AUC by over \revinsert{$9.8\%$}\revremove{$9.5\%$}, compared to Best-LR.  These results strongly suggest that all ITSs might benefit from logging the alternative features of AugmentedLR to improve the accuracy of their student performance modeling.

\input{tables/model_comparison}

\input{tables/num_params}

\subsection{The Cold Start Problem}
\label{subsec:cold_start}

\begin{figure}[!th]
    \centering
    \includegraphics[width=\textwidth]{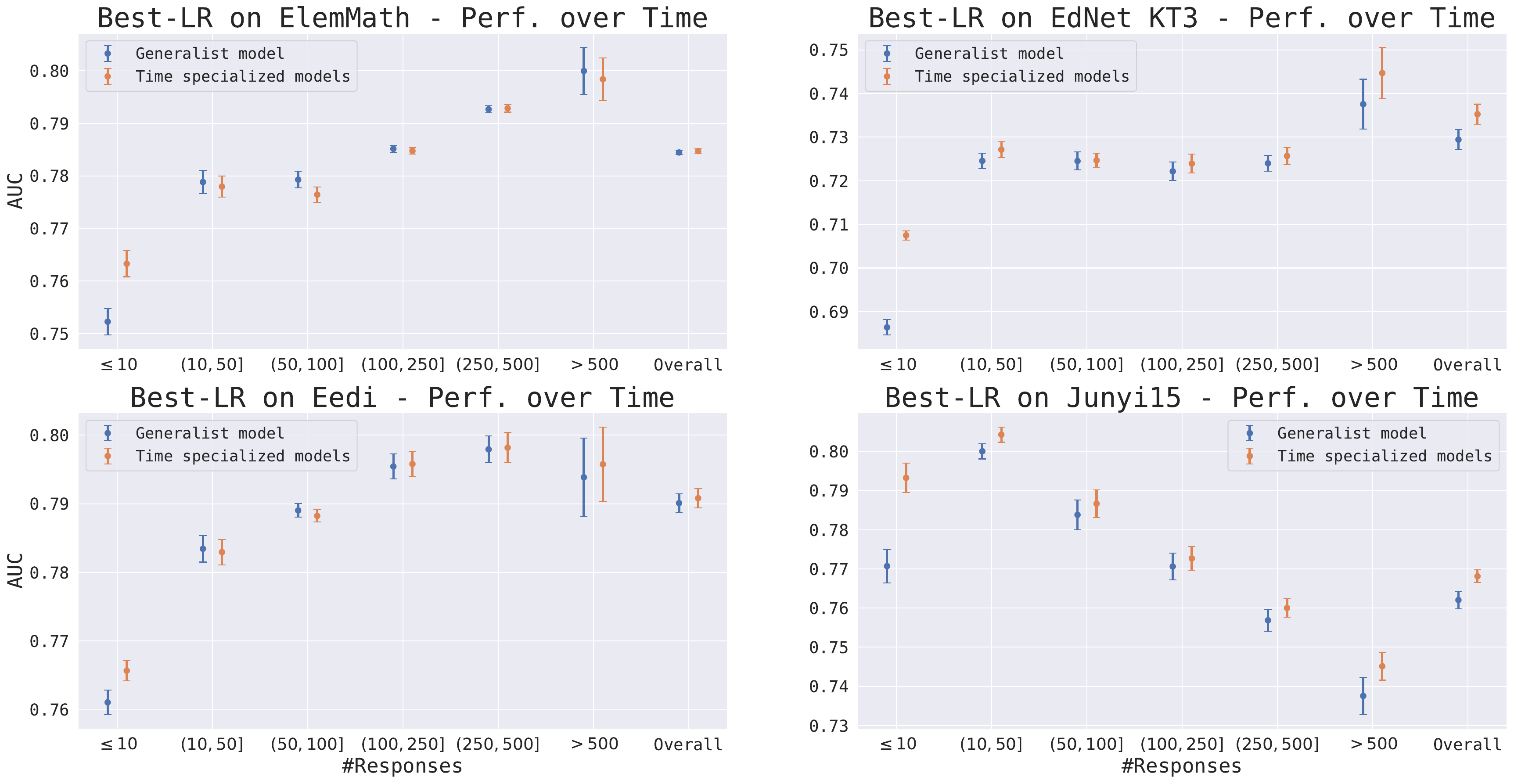}
    \caption{Comparison in AUC performance between a single Best-LR model trained using the entire training set (blue), and a collection of time-specialized Best-LR models trained on partitions of the response sequences (red). The horizontal axis shows the subset of question-response pairs used to train the specialized model. For example, the leftmost red point in each plot shows the AUC performance of a specialized model trained using only the first 10 question-response pairs of each student, then tested on the first 10 responses of other students. In contrast, the leftmost blue point shows the performance of the Best-LR model trained using all available data, and still tested on the first 10 responses of other students. The time-specialized models are able to mitigate the cold start problem for all four tutoring systems. On the \texttt{EdNet KT3} and \texttt{Junyi15} datasets the combined predictions of the specialized models increase overall performance substantially.}
    \label{fig:performance_over_time_bestlr}
\end{figure}

\begin{figure}[!th]
    \centering
    \includegraphics[width=\textwidth]{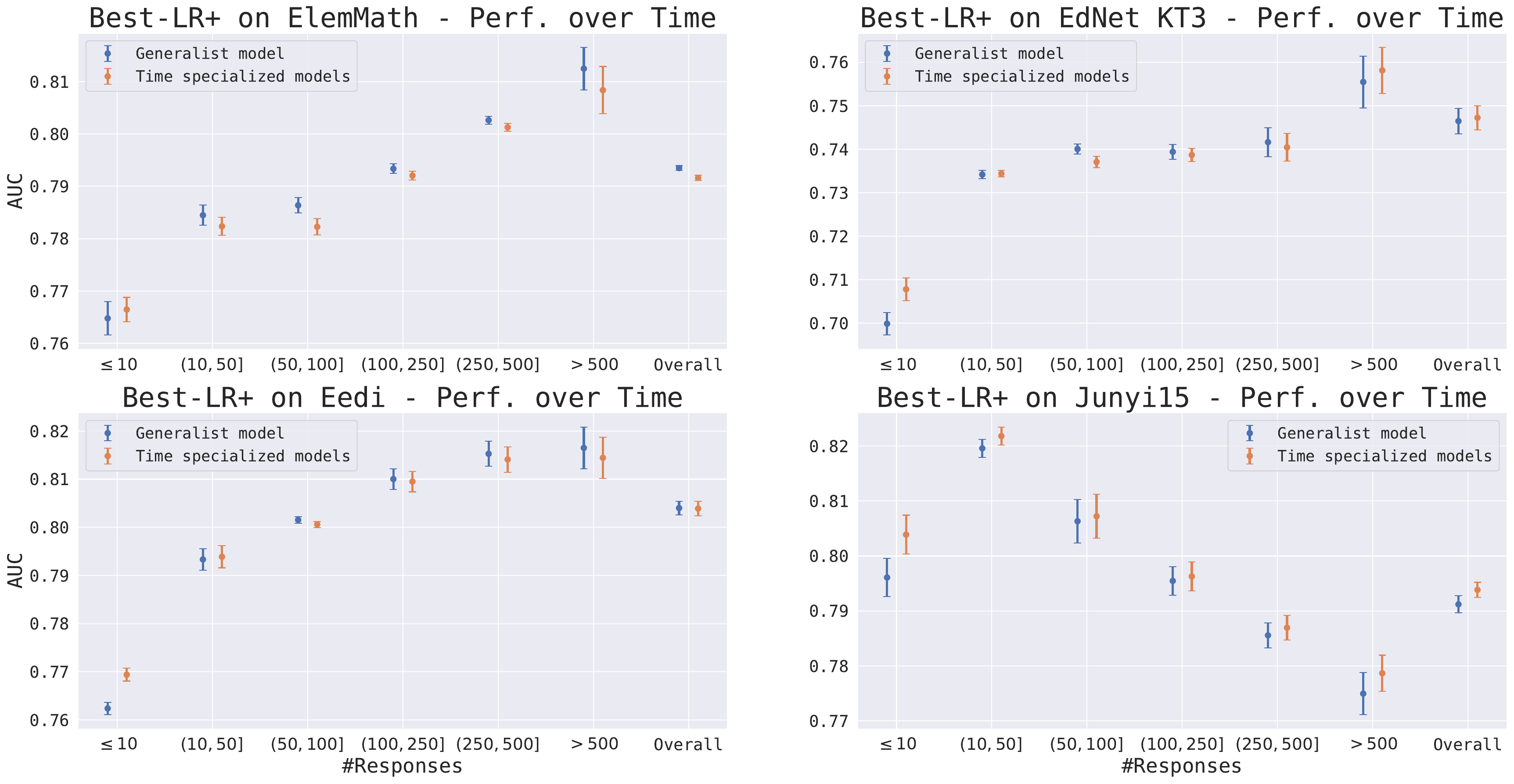}
    \caption{\revinsert{Comparison in AUC performance between a single Best-LR+ model trained using the entire training set (blue), and a collection of time-specialized Best-LR+ models trained on partitions of the response sequences (red). The horizontal axis shows the subset of question-response pairs used to train the specialized model. For example, the leftmost red point in each plot shows the AUC performance of a specialized model trained using only the first 10 question-response pairs of each student, then tested on the first 10 responses of other students. In contrast, the leftmost blue point shows the performance of the Best-LR+ model trained using all available data, and still tested on the first 10 responses of other students. The time-specialized models are able to mitigate the cold start problem for all four tutoring systems. On the \texttt{EdNet KT3} and \texttt{Junyi15} datasets the combined specialized predictions also improve overall performance.}}
    \label{fig:performance_over_time_bestlr_plus}
\end{figure}

Student performance modeling techniques use a student's interaction history to \revinsert{make predictions about their ability to solve different problems}\revremove{trace their knowledge state} over time. When a new student starts using the ITS, little or no interaction history is yet available, resulting in a new student "cold start" problem for estimating the \revinsert{performance}\revremove{knowledge state} of new students.  Most student performance models therefore require a burn-in period of student use of the ITS before they can accurately estimate student performance~(\citeNP{Gervet2020:Deep}, \citeNP{Zhang2021:Knowledge}). Here, to ensure a gratifying user experience and to improve early on retention rates, we show how one can mitigate the cold start problem by training multiple \emph{time-specialized} assessment models. We start by splitting the question-response sequence of each student into multiple distinct partitions based on their ordinal position in the student's learning process (i.e. partition \textit{50-100} will contain the 50th to 100th response of each students). We then train a separate time-specialized model for each individual partition. The motivation for this is that the way observational data needs to be interpreted can change over time. For example, during the beginning of the learning process one might put more focus on question and KC identifiers while later on count features provide a richer signal. With this approach a student's proficiency is evaluated by different specialized models depending on the length of their prior interactions history.

We evaluate \revinsert{the technique of}\revremove{this approach by} training \revinsert{multiple} time-specialized logistic regression models using the Best-LR \revinsert{and Best-LR+} feature set\revinsert{s using}\revremove{for each of} the four educational datasets \revinsert{(additional results for AugmentedLR are provided by Table~\ref{tab:augmented_composite_model})}. In particular we induce the partitions using the splitting points $\{ 0, 10, 50, 100, 250, 500, \infty \}$. Figure\revinsert{s}~\ref{fig:performance_over_time_bestlr} visualizes the results of a 5-fold cross-validation and \revinsert{compares predictive}\revremove{shows the} performance \revinsert{with}\revremove{of} a single \emph{generalist} Best-LR model trained using all available data. \revinsert{Figure~\ref{fig:performance_over_time_bestlr_plus} shows the corresponding experiment for Best-LR+. For both Best-LR and Best-LR+ feature sets} the use of time-specialized assessment models substantially improves prediction accuracy early on (i.e., during their first 10 question-response pairs) \revinsert{and mitigates the cold-start problem successfully} for all four datasets. For \texttt{EdNet KT3} and \texttt{Junyi} we observe AUC improvements of over $2\%$ \revinsert{and $0.8\%$} in early on predictions compared to the generalist \revinsert{Best-LR} \revinsert{and Best-LR+} model\revinsert{s respectively}. Also, for the same two datasets the overall \revinsert{Best-LR} performance (shown in Table~\ref{tab:composite_model}) achieved by combining the predictions of the individual time-specialized models yields an over $0.5\%$ increase in overall AUC scores and the time-specialized models consistently outperform the generalist model in each individual partition. For \texttt{ElemMath2021} and \texttt{Eedi} datasets the time-specialized \revinsert{Best-LR} models do not outperform the baseline consistently in each partition, but we still observe minor gains in predictive performance overall. \revinsert{Looking at the overall performance achieved by time-specialized Best-LR+ (Figure~\ref{fig:performance_over_time_bestlr_plus}) and time-specialized AugmentedLR models (Table~\ref{tab:augmented_composite_model}) we observe mixed results. While we observe consistent benefits for \texttt{EdNet KT3} and \texttt{Junyi15} the time-specialized models sometimes harm overall performance for \texttt{ElemMath2021} and \texttt{Eedi}. This might be due to the increased data intensity of the Best-LR+ and AugmentedLR models which we will discuss in more detail in Section~\ref{subsec:combining_multiple}.}

\subsection{Combining Multiple Specialized Models}
\label{subsec:combining_multiple}

\input{tables/multi_model}

\input{tables/informed_multi_model}

The large-scale datasets discussed in this paper are aggregations of learning trajectories collected from users of varying ages and grades who complete a range of different courses. The internal heterogeneity of the datasets paired with the large number of recorded interactions naturally leads to the question of whether it is more advantageous to pursue a monolithic or a composite modeling approach. Conventional student performance modeling techniques follow a monolithic approach in which they train a single model using all available training data, but the results in the previous section show that with sufficiently large datasets it can be useful to partition the training data to train multiple time-specialized models.
\revinsert{The idea of learning different classification rules for different parts of the dataset is a popular technique in the machine learning literature and a core principle behind decision tree algorithms~\cite{Breiman2001:Random}.
A related EDM question is whether it is more beneficial to model KCs in separation or in combination. For example while, BKT partitions the student interaction sequences by KC to infer KC-specific knowledge estimates, DKT takes in entire sequences to output a single vector which describes student proficiency for all KCs. Starting from this observation~\citeNP{Montero2018:Does} explore a modified DKT approach called DKT-SM-SS. Analogous to BKT, DKT-SM-SS partitions the student interaction sequences by KC and trains a separate KC-specific DKT model for each partition. Comparing the performance of DKT-SM-SS with conventional DKT they found that DKT benefits from modeling the KCs in combination. When partitioning the dataset per KC our logistic regression modeling approach has an advantage over DKT-SM-SS in that the underlying feature set still captures information about interactions with all KCs.}

In this section we explore the potential benefits of (1) training specialized models for specific questions, KCs, study modules, etc., and (2) combining the predictions from several of these models to obtain a final prediction of student performance.  The motivation for considering partitioning the data to train models specialized to these different contexts is that it has the potential to recover finer nuances in the training data, just as the time-specialized models in the previous section did. For example, two algebra courses might teach the same topics, but employ different analogies to promote conceptual understanding, allowing students to solve certain types of questions more easily than others. Training a separate specialized model for each course can allow the trained models to capture these differences and improve overall assessment quality.

Table~\ref{tab:composite_model} shows performance metrics for sets of logistic regression models trained using the Best-LR features. As baselines we use a single Best-LR model trained on the entire dataset as well as a set of time-specialized Best-LR models (described in Subsection~\ref{subsec:cold_start}). Note that even though question and KC identifiers are already part of the Best-LR feature set they can still be effective splitting criteria. As shown in the table, training question specialized models improves predictive performance for all datasets except \texttt{ElemMath2021} and KC specialized models are beneficial for all four datasets. Partitioning the data on the value of the study module feature, and training specialized models for each study module yields the greatest improvements for \texttt{ElemMath2021}, \texttt{EdNet KT3} and \texttt{Junyi15}, and also leads to substantial improvements on \texttt{Eedi}. It is interesting that training multiple specialized models for different study modules is more effective than augmenting the Best-LR feature set directly with the study module feature (Table~\ref{tab:bestlr_feature_performance}). Topic and course induced data partitions improve \revinsert{performance predictions}\revremove{knowledge assessments} for the \texttt{ElemMath2021} dataset. On the other hand, school and teacher specific splits are detrimental and we observe large overfitting to the training data. Splits on the bundle/quiz level are effective for \texttt{EdNet KT3} and \texttt{Eedi}. While \texttt{Eedi} benefits from incorporating the group identifiers into the Best-LR model, group specialized models harm overall performance.  

While fitting the question specific models on \texttt{ElemMath2021}, we observed severe overfitting behaviour, where accuracy on the training data is much higher than on the test set. This is likely caused by the fact that \texttt{ElemMath2021} contains the smallest number of responses per question among the four datasets. Table~\ref{tab:stats_composite_model} provides information about the average and median number of responses per model for the different data partitioning schemes.

We repeat the same experiment, but this time train logistic regression models using the ITS specific AugmentedLR feature sets. Performance metrics are provided by Table~\ref{tab:augmented_composite_model}. Unlike when using the Best-LR features, the question and KC specific AugmentedLR models have lower overall prediction quality than the original AugmentedLR model. These specialized AugmentedLR models contain many more parameters than their Best-LR counterparts due to the fact that they use more features to describe the student history (Table~\ref{tab:num_params}). This larger number of parameters requires a larger training dataset, which makes them more prone to overfitting. Still, splits on the study module, course and part features yield benefits on multiple datasets. Even though the AugmentedLR models benefit less from training specialized models, these specialized models still exhibit higher performance than the ones based on the Best-LR feature set.

\input{tables/stat_multi_model}

Finally, we discuss combining the strengths of the different models described in this section, by combining their predictions into a final group prediction \revinsert{via a machine learning technique called Stacking~\cite{Wolpert1992:Stacked}}. \revinsert{We do so by}\revremove{Our approach here is to} train\revinsert{ing} a higher-order logistic regression model that takes as input the predictions of the different models, and outputs the final predicted probability that the student will answer the question correctly. The learned weight parameters in this higher-order logistic regression model essentially create a weighted voting scheme that combines the predictions of the different models. We determined which models to include in this group prediction by evaluating all different combinations using the first data split of the 5-fold validation scheme and selecting the one yielding the highest AUC. The results of the best-performing combination models are shown in the final rows of Tables Table~\ref{tab:composite_model} (Combined-Best-LR) and Table~\ref{tab:augmented_composite_model} (Combined-AugmentedLR). Both tables also mark which models were used for this combination model, for each of the four datasets. 

The combination models outperform the baselines as well as the each individual set of specialized models for both Best-LR and AugmentedLR feature sets on all datasets. These combination models are the most accurate of all the logistic regression models discussed in this paper, and they also outperform all the deep learning baselines (Table~\ref{tab:model_comparison}) on all datasets except on \texttt{Eedi} where DKT is the only model that produces more accurate predictions. Looking at the Best-LR based combination models (Combined-Best-LR) we observe large AUC improvements of more than $0.65\%$ over the baseline models. For \texttt{EdNet KT3} and \texttt{Junyi15} there is an increase in AUC scores of $1.12\%$ and $6.09\%$ respectively. The minimum number of individual predictions used by a combination model is 3 (for \texttt{Eedi}) and the maximum number is 6 (for \texttt{EdNet KT3}). For the AugmentedLR based combination models (Combined-AugmentedLR) we observe AUC score improvements between \revinsert{$0.15\%$}\revremove{$0.16\%$} (for \revinsert{\texttt{Eedi}\revremove{\texttt{Junyi15}}}) and \revinsert{$0.47\%$}\revremove{$0.46\%$} (for \texttt{EdNet KT3}) compared to the baseline models. The AugmentedLR models already contain many of the features used for partitioning and are more data intensive then the Best-LR based models which is likely to be the reason for the smaller performance increment. The predictions of the combined AugmentedLR models rely mainly on the predictions of time- and study module-specialized models. Only the combination model for \revinsert{\texttt{ElemMath2021} uses course- instead of time-specialized models as its second signal.}\revremove{\texttt{Junyi15} uses the area-specialized models as a third signal.}

We note that while we were able to enhance prediction quality using only simple single feature partitioning schemes, future work on better strategies to partition the training data to train specialized models is likely to yield even larger benefits.

%% file: tables/individual_feature_evaluation.tex
\begin{table}[!t]
\scriptsize
\centering
\caption{Individual feature performance. Each entry reports ACC and AUC scores achieved by a logistic regression model trained using only the corresponding feature and a bias term. The first row provides a baseline for comparison, where no features are considered and the model simply predicts the student will answer every question correctly. Dashed lines indicate this feature is not available in this dataset. Maximum ACC and AUC variances over the five-fold test data are 0.17\% and 0.3\% respectively.}
\begin{tabular}{|l|cc|cc|cc|cc|}
\hline
& \multicolumn{2}{c|}{\texttt{ElemMath2021}} & \multicolumn{2}{c|}{\texttt{EdNet KT3}} & \multicolumn{2}{c|}{\texttt{Eedi}} & \multicolumn{2}{c|}{\texttt{Junyi15}}\\
Feature \,\textbackslash\, in \%    & ACC & AUC & ACC & AUC & ACC & AUC & ACC & AUC \\
\hline
always correct (baseline)  &             \avgvar{0.685246}{0.00083} & \avgvar{0.5}{0.0} & \avgvar{0.661862}{0.001721} & \avgvar{0.5}{0.0} & \avgvar{0.642973}{0.001304} & \avgvar{0.5}{0.0} & \avgvar{0.829885}{0.000535} & \avgvar{0.5}{0.0} \\
\hline
question ID      &                       \avgvar{0.723004}{0.000673} & \avgvar{0.728055}{0.000475} & \avgvar{0.694017}{0.001142} & \avgvar{0.703041}{0.001324} & \avgvar{0.672118}{0.000976} & \avgvar{0.685043}{0.000492} & \avgvar{0.831805}{0.000567} & \avgvar{0.720472}{0.001235} \\
KC ID      &                            \avgvar{0.696354}{0.000793} & \avgvar{0.666302}{0.000356}  &  \avgvar{0.662859}{0.000977} & \avgvar{0.588717}{0.000729} & \avgvar{0.642877}{0.001277} & \avgvar{0.574588}{0.001081} & \avgvar{0.830246}{0.00056} & \avgvar{0.640608}{0.001228} \\
\hline
total counts      &                      \avgvar{0.701424}{0.000928} & \avgvar{0.648614}{0.000857} &  \avgvar{0.664341}{0.000854} & \avgvar{0.595447}{0.002627} & \avgvar{0.690273}{0.001052} & \avgvar{0.720013}{0.001614} & \avgvar{0.832143}{0.000729} & \avgvar{0.646198}{0.001165} \\
KC counts      &                        \avgvar{0.707052}{0.000462} & \avgvar{0.66584}{0.000495} & \avgvar{0.661658}{0.000608} & \avgvar{0.602869}{0.001506} & \avgvar{0.683571}{0.000992} & \avgvar{0.697382}{0.001202} & \avgvar{0.837156}{0.000759} & \avgvar{0.688871}{0.001052} \\
question counts      &                        \avgvar{0.685249}{0.00083} & \avgvar{0.503224}{0.000255} & \avgvar{0.661887}{0.001025} & \avgvar{0.525428}{0.001417} & \avgvar{0.642973}{0.001304} & \avgvar{0.5}{0.0} & \avgvar{0.841906}{0.000707} & \avgvar{0.731108}{0.001683} \\
combined counts      &                        \avgvar{0.720527}{0.000476} & \avgvar{0.710517}{0.000543} & \avgvar{0.666484}{0.000816} & \avgvar{0.626332}{0.001838} & \avgvar{0.702751}{0.000975} & \avgvar{0.739355}{0.001448} & \avgvar{0.84346}{0.000713} & \avgvar{0.745701}{0.001235} \\
\hline
total TW counts                             & \avgvar{0.706406}{0.000606} & \avgvar{0.659615}{0.000953} & \avgvar{0.664847}{0.000855} & \avgvar{0.606181}{0.002554} & \avgvar{0.704195}{0.000909} & \avgvar{0.74322}{0.001362} & \avgvar{0.836009}{0.000671} & \avgvar{0.682676}{0.000432} \\
KC TW counts      &                        \avgvar{0.707182}{0.000463} & \avgvar{0.66757}{0.000391} & \avgvar{0.662524}{0.000663} & \avgvar{0.612179}{0.001391} & \avgvar{0.687767}{0.000941} & \avgvar{0.703534}{0.00114} & \avgvar{0.839872}{0.000624} & \avgvar{0.712269}{0.000535} \\
question TW counts      &                        \avgvar{0.685253}{0.000829} & \avgvar{0.503226}{0.000255} & \avgvar{0.661936}{0.000986} & \avgvar{0.530439}{0.001255} & \avgvar{0.642973}{0.001304} & \avgvar{0.5}{0.0} & \avgvar{0.842476}{0.000676} & \avgvar{0.738554}{0.000906} \\
combined TW counts      &                        \avgvar{0.723014}{0.000421} & \avgvar{0.71658}{0.000571} & \avgvar{0.667888}{0.000698} & \avgvar{0.640905}{0.002143} & \avgvar{0.71114}{0.000882} & \avgvar{0.751813}{0.001288} & \avgvar{0.843949}{0.00067} & \avgvar{0.754053}{0.000815} \\
\hline
R-PFA $F$ &                                               \avgvar{0.699457}{0.000581} & \avgvar{0.599677}{0.000449} & \avgvar{0.662175}{0.002927} & \avgvar{0.547179}{0.001472} & \avgvar{0.679232}{0.000973} & \avgvar{0.655482}{0.001306} & \avgvar{0.837441}{0.000791} & \avgvar{0.681462}{0.000555} \\
R-PFA $R$ &                                               \avgvar{0.685745}{0.000868} & \avgvar{0.635628}{0.000256} & \avgvar{0.661809}{0.003156} & \avgvar{0.602809}{0.001516} & \avgvar{0.642971}{0.001304} & \avgvar{0.619379}{0.000826} &  \avgvar{0.833673}{0.000554} & \avgvar{0.727109}{0.00057} \\
R-PFA $F$ + $R$ &                        \avgvar{0.704158}{0.000538} & \avgvar{0.665491}{0.000417} & \avgvar{0.663743}{0.002822} & \avgvar{0.616892}{0.001747} & \avgvar{0.686974}{0.000948} & \avgvar{0.703509}{0.001186} &  \avgvar{0.8416}{0.000714} & \avgvar{0.737755}{0.000463} \\
PPE Count &                                               \avgvar{0.692242}{0.000751} & \avgvar{0.634277}{0.00032} & \avgvar{0.661862}{0.003157} & \avgvar{0.590267}{0.001012} & \avgvar{0.64307}{0.001281} & \avgvar{0.564005}{0.001179} &  \avgvar{0.830329}{0.000538} & \avgvar{0.646585}{0.000733} \\
\hline
current elapsed time      &                        \avgvar{0.697176}{0.000839} & \avgvar{0.615027}{0.00099} & \avgvar{0.66181}{0.001708} & \avgvar{0.573552}{0.001393}  & - & - & \avgvar{0.832157}{0.000521} & \avgvar{0.662327}{0.002469} \\
current lag time      &                        \avgvar{0.685246}{0.00083} & \avgvar{0.515744}{0.000776}  & \avgvar{0.661862}{0.001721} & \avgvar{0.51783}{0.001715} & - & - & \avgvar{0.829885}{0.000535} & \avgvar{0.706948}{0.001181} \\
prior elapsed time  &                        \avgvar{0.694375}{0.000885} & \avgvar{0.558464}{0.000768} & \avgvar{0.661792}{0.001691} & \avgvar{0.524913}{0.001464} & - & - & \avgvar{0.829975}{0.000516} & \avgvar{0.601722}{0.001989} \\
prior lag time  &                        \avgvar{0.685246}{0.00083} & \avgvar{0.505782}{0.001076} & \avgvar{0.661862}{0.001721} & \avgvar{0.520032}{0.001589} & - & - &  \avgvar{0.829885}{0.000535} & \avgvar{0.537898}{0.000501} \\
\hline
month & \avgvar{0.685246}{0.00083} & \avgvar{0.513879}{0.000913}  & \avgvar{0.661887}{0.001025} & \avgvar{0.507262}{0.002213} & \avgvar{0.642973}{0.001304} & \avgvar{0.526248}{0.000645} & \avgvar{0.829885}{0.000535} & \avgvar{0.514351}{0.001711} \\
week & \avgvar{0.685246}{0.00083} & \avgvar{0.520594}{0.000879}  & \avgvar{0.661887}{0.001025} & \avgvar{0.507089}{0.002002} & \avgvar{0.642973}{0.001304} & \avgvar{0.530655}{0.000546} & \avgvar{0.829885}{0.000535} & \avgvar{0.516892}{0.001284} \\
day & \avgvar{0.685246}{0.00083} & \avgvar{0.505071}{0.000699}  &  \avgvar{0.661887}{0.001025} & \avgvar{0.502624}{0.00039} & \avgvar{0.642973}{0.001304} & \avgvar{0.515506}{0.000921} & \avgvar{0.829885}{0.000535} & \avgvar{0.512112}{0.000967} \\
hour & \avgvar{0.685246}{0.00083} & \avgvar{0.503644}{0.000712}  & \avgvar{0.661887}{0.001025} & \avgvar{0.506178}{0.000951} & \avgvar{0.642973}{0.001304} & \avgvar{0.516261}{0.000318} & \avgvar{0.829885}{0.000535} & \avgvar{0.521321}{0.001668} \\
\hline
study module ID &                        \avgvar{0.685246}{0.00083} & \avgvar{0.552764}{0.000334} &  \avgvar{0.662924}{0.001015} & \avgvar{0.544356}{0.001267} & \avgvar{0.646431}{0.000891} & \avgvar{0.583438}{0.001596} & \avgvar{0.829885}{0.000535} & \avgvar{0.663953}{0.001146} \\
study module counts &                        \avgvar{0.703542}{0.0005} & \avgvar{0.627643}{0.000745} & \avgvar{0.661836}{0.000824} & \avgvar{0.539977}{0.002618} & \avgvar{0.681271}{0.001087} & \avgvar{0.688202}{0.001398} & \avgvar{0.831735}{0.00085} & \avgvar{0.658533}{0.00078} \\
teacher/group      &                        \avgvar{0.683679}{0.000972} & \avgvar{0.565181}{0.000898} & - & -  & \avgvar{0.668462}{0.000925} & \avgvar{0.66843}{0.00144} & - & - \\
school       & \avgvar{0.68498}{0.001019} & \avgvar{0.556185}{0.00024} & - & - & - & - & - & - \\
course       & \avgvar{0.685258}{0.000825} & \avgvar{0.545593}{0.000933} & - & - & - & - & - & - \\
topic        & \avgvar{0.685395}{0.000844} & \avgvar{0.592656}{0.000568} & - & - & - & - & - & - \\
difficulty   & \avgvar{0.685511}{0.000821} & \avgvar{0.547256}{0.000126} & - & - & - & - & - & - \\
bundle/quiz      & -   & - & \avgvar{0.686971}{0.001135} & \avgvar{0.682099}{0.000999} & \avgvar{0.650461}{0.000973} & \avgvar{0.629856}{0.000772}  & - & - \\
part/area ID      & -   & - & \avgvar{0.661887}{0.001025} & \avgvar{0.56213}{0.000796} & - & - & \avgvar{0.830246}{0.00056} & \avgvar{0.570621}{0.000793} \\
part/area counts & - & - & \avgvar{0.665263}{0.000756} & \avgvar{0.610115}{0.00216} & - & - & \avgvar{0.834818}{0.000784} & \avgvar{0.661562}{0.000538} \\
\hline
age & - & - & - & - & \avgvar{0.642973}{0.001304} & \avgvar{0.533044}{0.001921} & - & - \\
gender & - & - & - & - & \avgvar{0.642973}{0.001304} & \avgvar{0.51505}{0.001753} & - & - \\
social support & - & - & - & - & \avgvar{0.642973}{0.001304} & \avgvar{0.551272}{0.0006} & - & - \\
platform     & \avgvar{0.685246}{0.00083} & \avgvar{0.501658}{0.000398} &  \avgvar{0.661887}{0.001025} & \avgvar{0.51734}{0.001058} & - & - & - & - \\
\hline
prereq IDs      &                        \avgvar{0.696358}{0.000792} & \avgvar{0.666299}{0.000355} & -   & - & \avgvar{0.64288}{0.001275} & \avgvar{0.574586}{0.001081} & \avgvar{0.831787}{0.000575} & \avgvar{0.720453}{0.00123} \\
prereq counts      &                        \avgvar{0.717348}{0.000477} & \avgvar{0.697169}{0.000538} & -   & - & \avgvar{0.692474}{0.000893} & \avgvar{0.718843}{0.001327} & \avgvar{0.844236}{0.000696} & \avgvar{0.759706}{0.001216} \\
postreq IDs      &                        \avgvar{0.696344}{0.000783} & \avgvar{0.666294}{0.000357} & - & - & - & - & \avgvar{0.831787}{0.000575} & \avgvar{0.720453}{0.00123} \\
postreq counts      & \avgvar{0.711237}{0.000535} & \avgvar{0.692542}{0.000263} & - & - & - & - & \avgvar{0.843388}{0.000736} & \avgvar{0.760268}{0.001089} \\
\hline
videos watched counts & \avgvar{0.685246}{0.00083} & \avgvar{0.534399}{0.000681} & \avgvar{0.661887}{0.001025} & \avgvar{0.542352}{0.002291} & - & - & - & - \\
videos skipped counts & \avgvar{0.68593}{0.000823} & \avgvar{0.572345}{0.000672}  & \avgvar{0.661887}{0.001025} & \avgvar{0.532654}{0.001685} & - & - & - & - \\
videos watched time & \avgvar{0.685243}{0.000829} & \avgvar{0.532872}{0.000541}  & \avgvar{0.661887}{0.001025} & \avgvar{0.540475}{0.001902} & - & - & - & - \\
\hline
reading counts & \avgvar{0.688967}{0.000874} & \avgvar{0.58916}{0.000772}  & \avgvar{0.661887}{0.001025} & \avgvar{0.552813}{0.002967} & - & - & - & - \\
reading time  & \avgvar{0.685023}{0.000851} & \avgvar{0.549245}{0.000592} & \avgvar{0.661887}{0.001025} & \avgvar{0.51475}{0.001835} & - & - & - & - \\
\hline
hint counts & - & - & - & - & - & - & \avgvar{0.829854}{0.000534} & \avgvar{0.599966}{0.001795} \\
hint time & - & - & - & - & - & - & \avgvar{0.829842}{0.000547} & \avgvar{0.597264}{0.002144} \\
\hline
smoothed avg correct &                        \avgvar{0.701843}{0.000878} & \avgvar{0.651422}{0.000863}
& \avgvar{0.664867}{0.000923} & \avgvar{0.598854}{0.002908} & \avgvar{0.691596}{0.001014} & \avgvar{0.720973}{0.001627} & \avgvar{0.829885}{0.000535} & \avgvar{0.663953}{0.001146} \\
response pattern  &                        \avgvar{0.708596}{0.000598} & \avgvar{0.649765}{0.001044}  & \avgvar{0.663898}{0.000661} & \avgvar{0.597539}{0.002392} & \avgvar{0.696663}{0.000851} & \avgvar{0.727113}{0.001298} & \avgvar{0.840195}{0.000619} & \avgvar{0.705053}{0.000234} \\
\hline
\end{tabular}
\label{tab:single_feature_performance}
\end{table}

%% file: tables/best_lr_feature_evaluation.tex
\begin{table}[!t]
\scriptsize
\centering
\caption{Augmented Best-LR performance. Each entry reports average ACC and AUC scores achieved by a logistic regression model trained using the Best-LR feature set augmented with a single feature. Maximum ACC and AUC variances over the five-fold test data are 0.15\% and 0.13\% respectively. The marker \no~is used to indicate features that are used for the AugmentedLR assessment models described later in the paper. Dashed lines indicate this feature is not available in this dataset.}
\begin{tabular}{|l|ccc|ccc|ccc|ccc|}
\hline
& \multicolumn{3}{c|}{\texttt{ElemMath2021}} & \multicolumn{3}{c|}{\texttt{EdNet KT3}} & \multicolumn{3}{c|}{\texttt{Eedi}} & \multicolumn{3}{c|}{\texttt{Junyi15}} \\
Feature \,\textbackslash\, in \%   & ACC & AUC & & ACC & AUC & & ACC & AUC & & ACC & AUC & \\
\hline
Best-LR (baseline) & \avgvar{0.75694}{0.000444} & \avgvar{0.784449}{0.000185} & \no & \avgvar{0.706875}{0.001438} & \avgvar{0.729425}{0.001155} & \no & \avgvar{0.734304}{0.000675} & \avgvar{0.790099}{0.000682} & \no & \avgvar{0.842537}{0.00079} & \avgvar{0.762043}{0.001122} & \no \\
\hline
question counts & \avgvar{0.756998}{0.000457} & \avgvar{0.784474}{0.000176} & & \avgvar{0.712394}{0.001065} & \avgvar{0.735335}{0.00059} & & \avgvar{0.734303}{0.000676} & \avgvar{0.7901}{0.000682} & & \avgvar{0.847369}{0.000707} & \avgvar{0.777551}{0.001177} &  \\
\hline
total TW counts      & \avgvar{0.759003}{0.000355} & \avgvar{0.787786}{0.000212} & & \avgvar{0.707487}{0.001488} & \avgvar{0.730733}{0.00126} &  & \avgvar{0.739405}{0.000705} & \avgvar{0.797277}{0.000637} & & \avgvar{0.843698}{0.000721} & \avgvar{0.766792}{0.000978} & \\
KC TW counts      & \avgvar{0.757248}{0.000429} & \avgvar{0.785429}{0.000159} & & \avgvar{0.709357}{0.001403} & \avgvar{0.732724}{0.001089} & & \avgvar{0.736361}{0.000608} & \avgvar{0.793016}{0.00063} & & \avgvar{0.844733}{0.000689} & \avgvar{0.769762}{0.000919} & \\
question TW counts      & \avgvar{0.757084}{0.000442} & \avgvar{0.78469}{0.000177} & & \avgvar{0.714149}{0.001109} & \avgvar{0.740503}{0.000681} & & \avgvar{0.734304}{0.000678} & \avgvar{0.7901}{0.000682} & & \avgvar{0.847805}{0.000731} & \avgvar{0.782655}{0.000994} & \\
combined TW counts      & \avgvar{0.759475}{0.000346} & \avgvar{0.788819}{0.000182} & \no & \avgvar{0.715059}{0.00118} & \avgvar{0.742307}{0.000539} & \no & \avgvar{0.73992}{0.000694} & \avgvar{0.797876}{0.000616} & \no & \avgvar{0.848354}{0.000674} & \avgvar{0.784008}{0.000937} & \no \\
\hline
R-PFA $F$ & \avgvar{0.757338}{0.000431} & \avgvar{0.785033}{0.000186} & & \avgvar{0.707662}{0.001738} & \avgvar{0.731377}{0.001104} & & \avgvar{0.737093}{0.000653} & \avgvar{0.793628}{0.000684} & & \avgvar{0.845558}{0.000727} & \avgvar{0.77226}{0.000862} & \\
R-PFA $R$ & \avgvar{0.757178}{0.000456} & \avgvar{0.784927}{0.000186} & & \avgvar{0.707482}{0.00174} & \avgvar{0.731}{0.001158} & & \avgvar{0.735659}{0.000691} & \avgvar{0.791508}{0.0007} & & \avgvar{0.847225}{0.000757} & \avgvar{0.778792}{0.000674} & \\
R-PFA $F$ + $R$ & \avgvar{0.75736}{0.000417} & \avgvar{0.785105}{0.00019} & \no & \avgvar{0.707989}{0.001758} & \avgvar{0.732485}{0.00118} & \no & \avgvar{0.737304}{0.000702} & \avgvar{0.793761}{0.000687} & \no & \avgvar{0.847713}{0.000731} & \avgvar{0.779483}{0.000676} & \no \\
PPE Count & \avgvar{0.757341}{0.00042} & \avgvar{0.785266}{0.000197} & \no & \avgvar{0.707334}{0.001719} & \avgvar{0.730736}{0.001151} & \no & \avgvar{0.734371}{0.000688} & \avgvar{0.790237}{0.000684} & \no & \avgvar{0.843389}{0.000786} & \avgvar{0.767466}{0.001047} & \no \\
\hline
current elapsed time      & \avgvar{0.760725}{0.000368} & \avgvar{0.789669}{0.000286} & & \avgvar{0.710121}{0.0013} & \avgvar{0.741422}{0.001045} & & - & - & & \avgvar{0.843773}{0.000704} & \avgvar{0.776249}{0.001142} & \\
current lag time & \avgvar{0.757899}{0.000404} & \avgvar{0.785384}{0.000188} & \no & \avgvar{0.707097}{0.001414} & \avgvar{0.730094}{0.001161} & \no & - & - & & \avgvar{0.843377}{0.000783} & \avgvar{0.76654}{0.001093} & \no \\
prior elapsed time  & \avgvar{0.758805}{0.000437} & \avgvar{0.786693}{0.000192} & \no & \avgvar{0.707739}{0.001381} & \avgvar{0.731078}{0.001132} & \no & - & - & & \avgvar{0.842636}{0.000795} & \avgvar{0.764225}{0.001037} & \no \\
prior lag time  & \avgvar{0.757556}{0.000429} & \avgvar{0.785228}{0.000197} & & \avgvar{0.707537}{0.001386} & \avgvar{0.730985}{0.001158} & & - & - & &  \avgvar{0.842602}{0.000779} & \avgvar{0.764539}{0.001059} & \\
\hline
month & \avgvar{0.756963}{0.000426} & \avgvar{0.784453}{0.000194} & & \avgvar{0.706893}{0.001435} & \avgvar{0.729483}{0.001159} & & \avgvar{0.734349}{0.000716} & \avgvar{0.790157}{0.000681} & & \avgvar{0.842546}{0.000772} & \avgvar{0.762073}{0.001115} & \\
week & \avgvar{0.757019}{0.000445} & \avgvar{0.784519}{0.000177} & & \avgvar{0.706897}{0.001438} & \avgvar{0.729473}{0.001167} & & \avgvar{0.734362}{0.000677} & \avgvar{0.790176}{0.00068} & & \avgvar{0.842553}{0.000786} & \avgvar{0.762141}{0.001127} & \\
day & \avgvar{0.756957}{0.00047} & \avgvar{0.784455}{0.000183} & & \avgvar{0.706859}{0.001418} & \avgvar{0.729431}{0.001151} & & \avgvar{0.73433}{0.000658} & \avgvar{0.790179}{0.000678} & & \avgvar{0.842632}{0.000811} & \avgvar{0.762444}{0.001129} & \\
hour & \avgvar{0.756956}{0.000464} & \avgvar{0.784466}{0.000182} & & \avgvar{0.706866}{0.001433} & \avgvar{0.729435}{0.001159} & & \avgvar{0.734355}{0.000666} & \avgvar{0.790143}{0.000678} & & \avgvar{0.842731}{0.000803} & \avgvar{0.762817}{0.001235} & \no \\
\hline
study module ID & \avgvar{0.757709}{0.000412} & \avgvar{0.785953}{0.000167} & \no & \avgvar{0.714043}{0.001324} & \avgvar{0.738845}{0.00109} & \no & \avgvar{0.735248}{0.000638} & \avgvar{0.791032}{0.000708} & \no & \avgvar{0.845598}{0.000753} & \avgvar{0.823897}{0.001023} & \no \\
study module counts & \avgvar{0.757405}{0.000408} & \avgvar{0.785441}{0.000174} & & \avgvar{0.707589}{0.001489} & \avgvar{0.730089}{0.001194} & & \avgvar{0.735187}{0.00072} & \avgvar{0.791031}{0.000746} & & \avgvar{0.842664}{0.000819} & \avgvar{0.763039}{0.001092} & \\
teacher/group      & \avgvar{0.756769}{0.000436} & \avgvar{0.783923}{0.000181} & & - & - & & \avgvar{0.740022}{0.000612} & \avgvar{0.796261}{0.000661} & \no & - & - & \\
school       & \avgvar{0.757186}{0.000431} & \avgvar{0.784751}{0.000189} & & - & - & & - & - & & - & - & \\
course       &  \avgvar{0.757177}{0.00043} & \avgvar{0.784869}{0.000191} & & - & - & & - & - & & - & - & \\
topic        & \avgvar{0.7574}{0.000441} & \avgvar{0.785329}{0.000207} & \no & - & - & & - & - & & - & - & \\
difficulty   & \avgvar{0.756973}{0.000455} & \avgvar{0.784495}{0.000184} & & - & - & & - & - & & - & - & \\
bundle/quiz ID       & - & - & & \avgvar{0.706868}{0.001429} & \avgvar{0.729416}{0.001154} & & \avgvar{0.737614}{0.000703} & \avgvar{0.794341}{0.000649} & \no & - & - & \\
part/area ID      & - & - & & \avgvar{0.706874}{0.001436} & \avgvar{0.729425}{0.001153} & & - & - & & \avgvar{0.842544}{0.000788} & \avgvar{0.76204}{0.00112} & \\
part/area counts & - & - & & \avgvar{0.707305}{0.001401} & \avgvar{0.730494}{0.001146} & \no & - & - & & \avgvar{0.842608}{0.000787} & \avgvar{0.762309}{0.001105} & \\
\hline
age & - & - & & - & - & & \avgvar{0.734477}{0.000654} & \avgvar{0.79026}{0.000687} & & - & - & \\
gender & - & - & & - & - & & \avgvar{0.734331}{0.00068} & \avgvar{0.790115}{0.000684} & & - & - & \\
social support & - & - & & - & - & & \avgvar{0.734358}{0.000662} & \avgvar{0.790182}{0.000674} & & - & - & \\
platform     & \avgvar{0.756962}{0.000449} & \avgvar{0.784457}{0.00018} & & \avgvar{0.706834}{0.00142} & \avgvar{0.729431}{0.001152} & & - & - & & - & - & \\
\hline
prereq IDs      & \avgvar{0.756954}{0.000443} & \avgvar{0.784451}{0.000189} & & - & - & & \avgvar{0.734301}{0.000675} & \avgvar{0.790101}{0.000683} & & \avgvar{0.84254}{0.000792} & \avgvar{0.762043}{0.00112} & \\
prereq counts      & \avgvar{0.759071}{0.000429} & \avgvar{0.787743}{0.000224} & \no & - & - & & \avgvar{0.735364}{0.00062} & \avgvar{0.791452}{0.000675} & \no & \avgvar{0.849139}{0.000742} & \avgvar{0.782031}{0.001044} & \no \\
postreq IDs      & \avgvar{0.756946}{0.000452} & \avgvar{0.784454}{0.000187} & & - & - & & - & - & & \avgvar{0.84254}{0.000792} & \avgvar{0.762043}{0.00112} & \\
postreq counts      & \avgvar{0.758148}{0.000436} & \avgvar{0.786387}{0.000169} & \no & - & - & & - & - & & \avgvar{0.848347}{0.000698} & \avgvar{0.780212}{0.000999} & \no \\
\hline
videos watched counts & \avgvar{0.757491}{0.000428} & \avgvar{0.785093}{0.000198} & \no & \avgvar{0.707023}{0.001389} & \avgvar{0.730397}{0.001152} & \no & - & - & & - & - & \\
videos skipped counts & \avgvar{0.757248}{0.000425} & \avgvar{0.784851}{0.000192} & & \avgvar{0.706966}{0.001362} & \avgvar{0.730012}{0.001117} & \no & - & - & & - & - & \\
videos watched time & \avgvar{0.7572}{0.000453} & \avgvar{0.784754}{0.000173} & & \avgvar{0.706841}{0.001414} & \avgvar{0.729853}{0.001182} & & - & - & & - & - & \\
\hline
reading counts & \avgvar{0.757481}{0.000405} & \avgvar{0.785828}{0.000179} & \no & \avgvar{0.706866}{0.00143} & \avgvar{0.729528}{0.001153} & & - & - & & - & - & \\
reading time  & \avgvar{0.756976}{0.000449} & \avgvar{0.784509}{0.000177} & & \avgvar{0.706933}{0.001405} & \avgvar{0.729566}{0.001074} & & - & - & & - & - & \\
\hline
hint counts & - & - & & - & - & & - & - & & \avgvar{0.843135}{0.00076} & \avgvar{0.765944}{0.00091} & \no \\
hint time & - & - & & - & - & & - & - & & \avgvar{0.842672}{0.000796} & \avgvar{0.763978}{0.001034} & \no \\
\hline
smoothed avg correct & \avgvar{0.757809}{0.000436} & \avgvar{0.786173}{0.000181} & \no & \avgvar{0.708064}{0.001306} & \avgvar{0.732168}{0.001091} & \no & \avgvar{0.734899}{0.000674} & \avgvar{0.791346}{0.000716} & \no & \avgvar{0.842755}{0.000802} & \avgvar{0.764247}{0.000968} & \no \\
response pattern  & \avgvar{0.760315}{0.00036} & \avgvar{0.789937}{0.000244} & \no & \avgvar{0.708216}{0.00139} & \avgvar{0.732445}{0.001128} & \no & \avgvar{0.743166}{0.000692} & \avgvar{0.800991}{0.000754} & \no & \avgvar{0.847211}{0.000651} & \avgvar{0.776504}{0.000821} & \no \\
\hline
\end{tabular}
\label{tab:bestlr_feature_performance}
\end{table}

%% file: tables/model_comparison.tex
\begin{table}[t]
\footnotesize
\centering
\caption{Comparative evaluation of student performance modeling algorithms across 4 large-scale datasets. The first four table rows correspond to previously studied logistic regression methods, the next four to previously studied deep neural network approaches, and the final two rows correspond to the two new logistic regression methods introduced in this paper.  Maximum ACC and AUC variances over the five-fold test data are 0.16\% and 0.17\% respectively. Because the \texttt{Eedi} dataset does not provide response time information it does not accommodate SAINT+ and there is no corresponding entry.}
\begin{tabular}{|l|cc|cc|cc|cc|}
\hline
& \multicolumn{2}{c|}{\texttt{ElemMath2021}} & \multicolumn{2}{c|}{\texttt{EdNet KT3}} & \multicolumn{2}{c|}{\texttt{Eedi}} & \multicolumn{2}{c|}{\texttt{Junyi15}} \\
Model \,\textbackslash\, in \%       & ACC & AUC & ACC & AUC & ACC & AUC & ACC & AUC \\
\hline
IRT      & \avgvar{0.723005}{0.000675} & \avgvar{0.728055}{0.000475} & \avgvar{0.694017}{0.001141} & \avgvar{0.703041}{0.001324} & \avgvar{0.672118}{0.000976} & \avgvar{0.685043}{0.000492} & \avgvar{0.831807}{0.000569} & \avgvar{0.720471}{0.001232} \\
PFA      & \avgvar{0.716845}{0.000478} & \avgvar{0.70796}{0.000415} & \avgvar{0.664783}{0.000637} & \avgvar{0.619519}{0.001285} & \avgvar{0.684386}{0.000928} & \avgvar{0.702794}{0.001087} & \avgvar{0.838525}{0.000739} & \avgvar{0.701949}{0.000842} \\
R-PFA & \avgvar{0.715985}{0.000561} & \avgvar{0.708036}{0.000407} & \avgvar{0.666002}{0.002644} & \avgvar{0.629195}{0.001601} & \avgvar{0.688968}{0.000922} & \avgvar{0.710115}{0.001109} & \avgvar{0.843494}{0.000757} & \avgvar{0.74129}{0.000442} \\
PPE & \avgvar{0.700843}{0.000754} & \avgvar{0.675907}{0.000417} & \avgvar{0.66509}{0.003075} & \avgvar{0.599299}{0.000856} & \avgvar{0.643243}{0.001295} & \avgvar{0.579496}{0.001067} & \avgvar{0.830897}{0.000597} & \avgvar{0.648068}{0.000814} \\
DAS3H      &  \avgvar{0.740822}{0.000459} & \avgvar{0.758162}{0.000271} & \avgvar{0.703127}{0.000661} & \avgvar{0.721598}{0.000809} & \avgvar{0.716444}{0.0007} & \avgvar{0.76132}{0.000412} & \avgvar{0.844307}{0.000679} & \avgvar{0.767253}{0.000947} \\
Best-LR      & \avgvar{0.75694}{0.000444} & \avgvar{0.784449}{0.000185} & \avgvar{0.706875}{0.001438} & \avgvar{0.729425}{0.001155} & \avgvar{0.734304}{0.000675} & \avgvar{0.790099}{0.000682} & \avgvar{0.842537}{0.00079} & \avgvar{0.762043}{0.001122} \\
\hline
DKT      & \avgvar{0.764618}{0.000290} & \avgvar{0.797148}{0.000349} & \avgvar{0.717671}{0.001515} & \avgvar{0.749240}{0.001215} & \textbf{\avgvar{0.754057}{0.000660}} & \textbf{\avgvar{0.815542}{0.000569}} & \avgvar{0.855025}{0.000763} & \avgvar{0.806150}{0.000738} \\
SAKT      & \avgvar{0.759004}{0.000472} & \avgvar{0.784432}{0.000485} & \avgvar{0.715294}{0.000275} & \avgvar{0.741055}{0.000417} & \avgvar{0.745066}{0.001085} & \avgvar{0.803051}{0.000792} & \avgvar{0.851623}{0.000659} & \avgvar{0.795885}{0.000638} \\
SAINT      & \avgvar{0.758673}{0.000161} & \avgvar{0.778769}{0.000191} & \avgvar{0.714033}{0.000469} & \avgvar{0.736917}{0.000581}
 & \avgvar{0.745638}{0.000599} & \avgvar{0.803753}{0.000509}
 & \avgvar{0.851007}{0.000411} & \avgvar{0.795071}{0.000997} \\
SAINT+ & \avgvar{0.760445}{0.000380} & \avgvar{0.781469}{0.000242} & \avgvar{0.715437}{0.000342} & \avgvar{0.739382}{0.000321} & - & - & \avgvar{0.851796}{0.000460} & \avgvar{0.797114}{0.001108} \\
\hline
Best-LR+    & \avgvar{0.76227}{0.000325} & \avgvar{0.793474}{0.00023} &  \avgvar{0.716949}{0.001991} & \avgvar{0.746452}{0.001465} & \avgvar{0.745483}{0.000666} & \avgvar{0.803994}{0.000715} & \avgvar{0.850471}{0.000679} & \avgvar{0.791201}{0.000769} \\
AugmentedLR    & \textbf{\avgvar{0.765883}{0.000307}} & \textbf{\avgvar{0.798676}{0.000262}} & \textbf{\avgvar{0.718932}{0.001907}} & \textbf{\avgvar{0.749953}{0.001319}} & \avgvar{0.749568}{0.000706} & \avgvar{0.809607}{0.000696} & \textbf{\avgvar{0.86348}{0.000523}} & \textbf{\avgvar{0.860258}{0.000533}} \\
\hline
\end{tabular}
\label{tab:model_comparison}
\end{table}

%% file: tables/num_params.tex
\begin{table}[!t]
\footnotesize
\centering
\caption{ Number of parameters learned by different student performance models. The number of parameters is heavily dependent on the number of questions and KCs used by the underlying ITS.}
\begin{tabular}{|l|c|c|c|c|}
\hline
& \texttt{ElemMath2021} & \texttt{EdNet KT3} & \texttt{Eedi} & \texttt{Junyi15} \\
\hline
\# of unique questions & 59,892 & 13,169 & 27,613 & 835 \\
\# of KCs & 4,191 & 293 & 388 & 41 \\
\hline
IRT & 59,571 & 11,556 & 27,614 & 723 \\
PFA & 12,574 & 904 & 1,165 & 124 \\
R-PFA & 12,574 & 904 & 1,165 & 124 \\
PPE & 8,383 & 603 & 777 & 83 \\
DAS3H & 105,672 & 14,867 & 31,882 & 1,174 \\
Best-LR & 72,146 & 12,461 & 28,780 & 848 \\
\hline
DKT & 21,529,751 & 526,526 & 10,676,751 & 184,801 \\
SAKT & 6,730,901 & 1,033,051 & 3,191,301 & 125,101 \\
SAINT & 16,727,873 & 8,026,049 & 7,666,497 & 1,618,049 \\
SAINT+ & 4,194,241 & 1,349,889 & - & 148,865 \\
\hline
Best-LR+ & 119,291 & 16,816 & 34,092 & 2,343 \\
AugmentedLR & 137,602 & 17,297 & 64,076 & 6,170 \\
Combined-AugmentedLR & 10,595,355 & 242,159 & 4,164,941 & 86,381 \\
\hline
\end{tabular}
\label{tab:num_params}
\end{table}

%% file: tables/multi_model.tex
\begin{table}[!t]
\scriptsize
\centering
\caption{Composite Best-LR performance for different partitioning schemes. The first two rows of the table give as baseline scores the ACC and AUC for the previously discussed Best-LR model and time-specialized Best-LR models (Subsection~\ref{subsec:cold_start}). The next 10 rows show results when training specialized models that partition the data based on single features such as question ID, KC ID, etc..  The final row shows the result of combining several of these models by taking a weighted vote of their predictions as described in the text. The marker \no~indicates the inputs used for this combination model for each dataset. Maximum ACC and AUC variances over the five-fold test data are 0.14\% and 0.12\% respectively.}
\begin{tabular}{|l|ccc|ccc|ccc|ccc|}
\hline
& \multicolumn{3}{c|}{\texttt{ElemMath2021}} & \multicolumn{3}{c|}{\texttt{EdNet KT3}} & \multicolumn{3}{c|}{\texttt{Eedi}} & \multicolumn{3}{c|}{\texttt{Junyi15}} \\
Feature \,\textbackslash\, in \% & ACC & AUC & & ACC & AUC & & ACC & AUC & & ACC & AUC & \\
\hline
Best-LR & \avgvar{0.756943}{0.000453} & \avgvar{0.784452}{0.000182} & & \avgvar{0.706875}{0.001438} & \avgvar{0.729425}{0.001155} & \no & \avgvar{0.734304}{0.000675} & \avgvar{0.790099}{0.000682} & & \avgvar{0.842537}{0.00079} & \avgvar{0.762043}{0.001122} & \no \\
Best-LR (time-spec.) & \avgvar{0.757138}{0.000501} & \avgvar{0.784715}{0.000219} & & \avgvar{0.710443}{0.001356} & \avgvar{0.735263}{0.001137} & \no & \avgvar{0.735097}{0.000734} & \avgvar{0.790813}{0.000706} & \no & \avgvar{0.843311}{0.000823} & \avgvar{0.768129}{0.000827} & \no \\
\hline
question ID specific & \avgvar{0.757288}{0.00045} & \avgvar{0.783121}{0.000265} & & \avgvar{0.709023}{0.001257} & \avgvar{0.731776}{0.001046} & & \avgvar{0.736938}{0.000678} & \avgvar{0.791203}{0.000677} & \no & \avgvar{0.843747}{0.000727} & \avgvar{0.767561}{0.000886} & \no \\
KC ID specific & \avgvar{0.758094}{0.000425} & \avgvar{0.786277}{0.0002} & \no & \avgvar{0.708004}{0.001409} & \avgvar{0.73108}{0.001133} & \no & \avgvar{0.734972}{0.000689} & \avgvar{0.790768}{0.000706} & & \avgvar{0.842593}{0.000777} & \avgvar{0.762646}{0.001088} & \\
study module specific & \avgvar{0.759575}{0.000395} & \avgvar{0.789433}{0.000228} & \no & \avgvar{0.716478}{0.001434} & \avgvar{0.744324}{0.001214} & \no & \avgvar{0.737279}{0.00064} & \avgvar{0.793005}{0.000697} & & \avgvar{0.846866}{0.000823} & \avgvar{0.826747}{0.000957} & \no \\
teacher/group specific & \avgvar{0.706197}{0.000534} & \avgvar{0.669142}{0.000602} & & - & - & & \avgvar{0.729894}{0.000645} & \avgvar{0.781455}{0.001042} & & - & - & \\
school specific & \avgvar{0.72174}{0.000545} & \avgvar{0.713292}{0.000572} & & - & - & & - & - & & - & - & \\
course specific & \avgvar{0.757891}{0.000427} & \avgvar{0.785895}{0.00019} & \no & - & - & & - & - & & - & - & \\
topic specific & \avgvar{0.758249}{0.000147} & \avgvar{0.786817}{0.000148} & & - & - & & - & - & & - & - & \\
bundle/quiz specific & - & - & & \avgvar{0.709216}{0.001261} & \avgvar{0.732412}{0.001036} & \no & \avgvar{0.739481}{0.000692} & \avgvar{0.794855}{0.000707} & \no & - & - & \\
part/area specific & - & - & & \avgvar{0.707103}{0.001394} & \avgvar{0.730113}{0.001098} & & - & - & & \avgvar{0.842547}{0.000777} & \avgvar{0.762276}{0.001111} & \no \\
platform specific & \avgvar{0.756683}{0.000471} & \avgvar{0.783869}{0.000196} & \no & \avgvar{0.70707}{0.001383} & \avgvar{0.729779}{0.001008} & \no & - & - & & - & - & \\
\hline
Combined-Best-LR & \avgvar{0.760781}{0.000381} & \avgvar{0.791339}{0.000205} & & \avgvar{0.717533}{0.001277} & \avgvar{0.746496}{0.001098} & & \avgvar{0.741295}{0.000716} & \avgvar{0.798257}{0.000688} & & \avgvar{0.847052}{0.000822} & \avgvar{0.828982}{0.000816} & \\
\hline
\end{tabular}
\label{tab:composite_model}
\end{table}

%% file: tables/informed_multi_model.tex
\begin{table}[!t]
\scriptsize
\centering
\caption{Composite AugmentedLR performance for different partitioning schemes. The first two rows of the table give as baseline scores the ACC and AUC for the previously discussed AugmentedLR model and time-specialized AugmentedLR models (Subsection~\ref{subsec:cold_start}). The next 10 rows show results when training specialized models that partition the data based on single features such as question ID, KC ID, etc.. The final row shows the result of combining several of these models by taking a weighted vote of their predictions as described in the text. The marker \no~indicates the inputs used for this combination model for each dataset.  Maximum ACC and AUC variances over the five-fold test data are 0.19\% and 0.14\% respectively.}
\begin{tabularx}{\textwidth}{|X|ccc|ccc|ccc|ccc|}
\hline
& \multicolumn{3}{c|}{\texttt{ElemMath2021}} & \multicolumn{3}{c|}{\texttt{EdNet KT3}} & \multicolumn{3}{c|}{\texttt{Eedi}} & \multicolumn{3}{c|}{\texttt{Junyi15}} \\
Feature \,\textbackslash\, in \% & ACC & AUC & & ACC & AUC & & ACC & AUC & & ACC & AUC & \\
\hline
AugmentedLR &                            \avgvar{0.765898}{0.000299} & \avgvar{0.79869}{0.00027}  & & \avgvar{0.718934}{0.001898} & \avgvar{0.749949}{0.00132} & & \avgvar{0.749576}{0.000707} & \avgvar{0.809607}{0.000697} & & \avgvar{0.863662}{0.000519} & \avgvar{0.860346}{0.000527} & \\
AugmentedLR (time-spec.) &                \avgvar{0.76406}{0.000308} & \avgvar{0.795106}{0.000289}  & &  \avgvar{0.71917}{0.001852} & \avgvar{0.750054}{0.001422}  & \no & \avgvar{0.748342}{0.000693} & \avgvar{0.808004}{0.000752} & \no & \avgvar{0.863762}{0.000653} & \avgvar{0.861626}{0.000583} & \no \\
\hline
question ID specific &                   \avgvar{0.750947}{0.000355} & \avgvar{0.773717}{0.000393} & & \avgvar{0.701567}{0.001779} & \avgvar{0.723579}{0.00108} & & \avgvar{0.74334}{0.000744} & \avgvar{0.801353}{0.000761} & & \avgvar{0.861165}{0.000479} & \avgvar{0.857156}{0.000446} & \\
KC ID specific &                           \avgvar{0.757299}{0.000333} & \avgvar{0.784933}{0.000295} & & \avgvar{0.712136}{0.001917} & \avgvar{0.738598}{0.001384} & & \avgvar{0.745907}{0.000746} & \avgvar{0.805655}{0.000778} & & \avgvar{0.861573}{0.000647} & \avgvar{0.858229}{0.000608} & \\
study module specific &                  \avgvar{0.767052}{0.000262} & \avgvar{0.80031}{0.000306} & \no & \avgvar{0.720177}{0.00169} & \avgvar{0.753417}{0.001308} & \no & \avgvar{0.749232}{0.000664} & \avgvar{0.809286}{0.000717} & \no & \avgvar{0.864172}{0.000611} & \avgvar{0.861912}{0.000632} & \no \\
teacher/group specific & \avgvar{0.669584}{0.000651} & \avgvar{0.611563}{0.00113} & & - & - & & \avgvar{0.725268}{0.000773} & \avgvar{0.776181}{0.001037} & & - & - & \\
school specific & \avgvar{0.700274}{0.000727} & \avgvar{0.668237}{0.000867} & & - & - & & - & - & & - & - & \\
course specific & \avgvar{0.764925}{0.000307} & \avgvar{0.796936}{0.000262} & \no & - & - & & - & - & & - & - & \\
topic specific & \avgvar{0.759843}{0.000311} & \avgvar{0.788858}{0.000372} & & - & - & & - & - & & - & - & \\
bundle/quiz specific & - & - & & \avgvar{0.702156}{0.001777} & \avgvar{0.724379}{0.001121} & & \avgvar{0.739558}{0.000798} & \avgvar{0.796484}{0.000785} & & - & - & \\
part/area specific & - & - & & \avgvar{0.719106}{0.001856} & \avgvar{0.750853}{0.001297} & & - & - & & \avgvar{0.862942}{0.000566} & \avgvar{0.859891}{0.000525} & \\
platform & \avgvar{0.765489}{0.000293} & \avgvar{0.79778}{0.000294}  & & \avgvar{0.719151}{0.001724} & \avgvar{0.750417}{0.001334} & & - & - & & - & - & \\
\hline
Combined-AugmentedLR & \avgvar{0.767581}{0.000273} & \avgvar{0.801615}{0.000272} & & \avgvar{0.721139}{0.001735} & \avgvar{0.75479}{0.001386} & & \avgvar{0.750434}{0.000737} & \avgvar{0.811095}{0.000729} & & \avgvar{0.864583}{0.000641} & \avgvar{0.863374}{0.000603} & \\
\hline
\end{tabularx}
\label{tab:augmented_composite_model}
\end{table}

%% file: tables/stat_multi_model.tex
\begin{table}[!t]
\scriptsize
\centering
\caption{Average and median number of training examples (i.e. question responses) available for different partitioning schemes. For example, when a specialized model is trained for each different question ID in the \texttt{ElemMath2021} dataset, the average number of training examples available per question-specific model is 393.}
\begin{tabular}{|l|cc|cc|cc|cc|}
\hline
& \multicolumn{2}{c|}{\texttt{ElemMath2021}} & \multicolumn{2}{c|}{\texttt{EdNet KT3}} & \multicolumn{2}{c|}{\texttt{Eedi}} & \multicolumn{2}{c|}{\texttt{Junyi15}} \\
Feature & Avg & Med & Avg & Med & Avg & Med & Avg & Med \\
\hline
question ID specific & 393 & 118 & 1,445 & 930 & 718 & 351 & 35,207 & 8,380 \\
KC ID specific & 5,801 & 2,031 & 11,694 & 2,956 & 18,485 & 613 & 635,487 & 200,284 \\
study module specific & 3,901,401 & 2,487,865 & 2,385,616 & 810,120 & 336,183 & 29,589 & 5,083,895 & 3,421,938 \\
teacher/group specific & 1,370 & 297 & - & - & 1,674 & 554 & - & - \\
school specific & 11,992 & 5,620 & - & - & - & - & - & - \\
course specific & 329,695 & 69,125 & - & - & - & - & - & - \\
topic specific & 21,614 &  4809 & - & - & - & - & - & - \\
bundle/quiz specific & - & - & 1,970 & 1,265 & 1,146 & 120 & - & - \\
part/area specific & - & - & 2,385,616 & 1,344,293 & - & - & 2,824,386 & 605,666 \\
platform & 11,704,204 & 11,704,204 & 8,349,656 & 8,349,656 & - & - & - & - \\
\hline
\end{tabular}
\label{tab:stats_composite_model}
\end{table}

%% file: text/discussion.tex
\section{Results Summary and Discussion}
\label{sec:discussion}

The main results of this paper show that the state of the art of student performance modeling can be advanced through new machine learning approaches, yielding more accurate student assessments and in particular more accurate predictions of which questions a student will be able to answer correctly at any given point in time as they move through the online course. In particular we show:
\begin{itemize}
    \item State-of-the-art logistic regression approaches to student performance modeling can be further improved by incorporating a set of {\em new features that can be easily calculated from the question-response pairs} that appear in student log data from nearly all ITSs. For example, these include counts and smoothed ratios of correctly answered questions and overall attempts, partitioned by question, by knowledge component, and by time window, as well as specific sequences of correct/incorrect responses over the most recent student responses. We refer to the previous state-of-the-art logistic regression model as Best-LR~\cite{Gervet2020:Deep} and to the logistic regression model that incorporates these additional features as Best-LR+. Our experiments show that Best-LR+ yields more accurate student modeling than Best-LR across all four diverse ITS logs considered in this paper. We conclude that most tutoring systems that perform student modeling should benefit by incorporating these features.
    
    \item A second way of improving over the state of the art in student modeling is to incorporate {\em new types of features that go beyond the traditional question-response data} typically logged in all ITSs. For example, accuracy is improved by incorporating features such as the time students took to answer the previous questions, student performance on earlier questions associated with prerequisites to the knowledge component of the current question, and information about the study module (e.g., does the question appear in a pre-test, post-test, or as a practice problem). We conclude that future tutoring systems should log the information needed to provide these features to their student performance modeling algorithms.
    
    \item A third way to improve of the state of the art is to train {\em multiple, specialized student performance models} and then combine their predictions to form a final group prediction. For example, we found that training distinct logistic regression models on different partitions of the data (e.g., partitioning the data by its position in the sequential log, or by the knowledge component being tested) leads to improved accuracy. Furthermore, combining the predictions of different specialized models leads to additional accuracy improvements (e.g., combining the predictions of specialized models trained on different question bundles, with predictions of specialized models trained on different periods of time in the sequential log). We conclude that time-specialized models can help ameliorate the problem of assessing new students who have not yet created a long sequence of log data.  Furthermore, we feel that as future tutoring systems are adopted by more and more students, the increasing size of student log datasets will make this approach of training and combining specialized models increasingly effective.
    
    \item Although our primary focus here is on logistic regression models, we also considered top-performing neural network approaches including DKT~\cite{Piech2015:Deep}, SAKT~\cite{Pandey2019:Self}, SAINT~\cite{Choi2020:Towards} and SAINT+~\cite{Shin2021:Saint+} as additional state-of-the-art systems against which we compare. Our experiments show that among these neural network approaches, DKT consistently outperforms the others across our four diverse datasets. However, we also find that our logistic regression model Combined-AugmentedLR, which combines the three above points, outperforms all of these neural network models on average across the four datasets, and outperforms them all on three of the four individual datasets (DKT outperforms Combined-AugmentedLR on the \texttt{Eedi} dataset). We do find that neural network approaches are promising, however, especially due to their ability in principle to automatically discover additional features that logistic regression cannot discover on its own. Furthermore, we believe this ability of neural networks will be increasingly important as available student log datasets continue to increase both in size and in diversity of logged features. We conclude that a promising direction for future research is to explore the integration of our above three approaches into DKT and other neural network approaches.
\end{itemize}

It is useful to consider how our results relate to previously published results. We found that the time window features proposed by DAS3H~\cite{Choffin2019:DAS3H} enhanced Best-LR assessments for all four datasets, providing strong support for their proposed features. Note that when~\citeNP{Gervet2020:Deep} introduced the Best-LR model they also experimented with time window features, but unlike us did not observe consistent benefits. This might be due to the number of additional parameters that must be trained when incorporating these time window features, and the corresponding need for larger training datasets. Recall the datasets we used in this manuscript are about one order of magnitude larger than the ones used by~\citeNP{Gervet2020:Deep}.
\revinsert{Our algorithm comparison (Table~\ref{tab:model_comparison}) revealed that the prediction quality of PPE~\cite{Walsh2018:Mechanisms} is worse than all other considered student performance modeling techniques. This is likely due to the fact that PPE was designed to model cognitive processes during word pair learning over longer periods of time. This is a very different setting then the learning experiences offered by the four studied tutoring systems. Three of them focus on mathematics and \texttt{EdNet KT3} prepares students for the TOEIC\textsuperscript{\textcopyright} examination which goes beyond conventional retrieval practice.}
The elapsed and lag time features introduced by SAINT+~\cite{Shin2021:Saint+} also improve Best-LR predictions substantially in our experiments. Interestingly, the performance increment for the Best-LR model produced by these features is comparable to and sometimes even greater than the performance difference between SAINT and SAINT+. Note SAINT+ is an improved version of SAINT that uses the two interaction time based features. This suggests it might not require a deep learning-based approach to leverage these elapsed time and lag time features optimally.

Considering features that require augmented student logs that contain more than question-response pairs, we found these augmented features vary in utility. We found the feature "current study module" to be a particularly useful signal for all datasets. During preliminary analysis we observed differences in student performance between different parts of the learning sessions (i.e. pre-test, learning, post-test, \dots), which are captured by the study module feature. Even though post-tests tend to contain more difficult questions on average, the highest level of student performance is observed during post-test session in which the users' overall performance is evaluated.

Importantly, we also found that introducing background knowledge about prerequisites and postrequisites among the knowledge components (KCs) in the curriculum is very useful. As summarized in Table~\ref{tab:single_feature_performance}, counts of correctly answered questions and attempts associated with pre- and post-requisite KCs are among the most informative features. Importantly, these features can be easily incorporated even into pre-existing ITS's and datasets that log only question-response pairs, because calculating these features requires no new log data -- only annotating it using background knowledge about their prerequisite structure.

We compared different types of machine learning algorithms for student performance modeling in Subsection~\ref{subsec:integrating_features}. One limitation is that we did not refit IRT's student ability parameter after each user response. \citeNP{Wilson2016:Estimating} showed that refitting the ability parameter after each interaction makes IRT more competitive on multiple smaller datasets. \revinsert{While our feature selection for the AugmentedLR models solely focused on achieving accurate performance predictions, the inclusion of certain contextual features can lead to reductions in generalizability to unseen data. For example, the use of school or teacher specific parameters requires us to refit the model periodically as new schools and teachers  start working with the system. When selecting a feature set for real world applications one might want to trade a small reduction in predictive performance in favor of enhanced generalizability to new users.}

Recently, multiple intricate deep learning based techniques have been proposed and yield state-of-the-art performance for specific datasets~(e.g. \citeNP{Shin2021:Saint+},~\citeNP{Zhou2021:Lana},~\citeNP{Zhang2021:Muse}, \dots). Unfortunately, many of these works only employ one or two datasets which raises the question of how suitable they are for other tutoring systems. The code and new data released alongside this manuscript increases the usability of multiple large-scale educational datasets. We hope that future works will leverage these available datasets to test whether novel student performance modeling algorithms are effective across different tutoring systems.

%% file: text/conclusion.tex
\section{Conclusion}
\label{sec:conclusion}

In this paper we present several approaches to extending the state of the art in student performance modeling.  We show that approaches based on logistic regression can be improved by adding specific new features to student log data, including features that can be calculated from simple question-response student logs, and features not traditionally captured in student logs such as which videos the student watched or skipped.  Furthermore, we introduce a new approach of training multiple logistic regression models on different partitions of the data, and show further improvements in student performance modeling by automatically combining the predictions of these multiple specialized logistic regression models.

Taken together, our proposed improvements lead to our Combined-AugmentedLR method which achieves a new state of the art for student performance modeling. Whereas the previous state-of-the-art logistic regression approach, Best-LR~\cite{Gervet2020:Deep}, achieves an AUC score of 0.767 on average over our four datasets, our Combined-AugmentedLR approach achieves an improved AUC of 0.808 -- a reduction of 17.5\% in the AUC error (i.e., in the difference between the achieved AUC score and the ideal AUC score of 1.0).  Furthermore, we observed that Combined-AugmentedLR achieves improvements consistently across all four of the diverse datasets we considered, suggesting that our methods should be useful across a broad range of intelligent tutoring systems. 

To encourage researchers to compare against our approach, three of the four datasets we chose are publicly available.\revremove{We make the fourth, \texttt{Squirrel Ai ElemMath2021}, available via Carnegie Mellon University's DataShop.}
In addition, we make the entire code base for the algorithms and experiments reported here available on GitHub\footnote{\url{https://github.com/rschmucker/Large-Scale-Knowledge-Tracing}}.  Our implementation converts the four large-scale datasets into a standardized format and uses parallelization for efficient processing. It increases the usability of the large-scale log data, and we hope that it will benefit future research on novel student performance modeling techniques.

%% file: text/appendix.tex
\section{Appendix: Extended Feature Description}
\label{app:features}

This Appendix provides a reference with additional implementation details for the features described in Section~\ref{sec:approach}. The individual ITS capture similar information in different ways and we discuss necessary system specific adaptions. While we are already sharing the complete code base that was used to generate the experimental results in this paper on GitHub, this Appendix is intended to guide independent re-implementations.

\subsection*{Question (Item) ID}

All four of our datasets assign each question (i.e., item) a unique  identifier. To make a performance prediction for a question our implementation converts the corresponding question identifier into a sparse one-hot vector which is then passed to the machine learning algorithm.  Note a one-hot vector refers to a vector where all values are zero except for a single value of 1.  For example, given a dataset with $n$ different questions, our one-hot vector representation contains $n$ values, of which $n-1$ are zeros, and just a single value of 1 to indicate the current question. Knowledge about question identifiers allows a model to learn question specific difficulty parameters. 

\subsection*{Knowledge Component (KC) ID}

Knowledge about KC identifiers allows a model to learn skill specific difficulty parameters. While \texttt{ElemMath2021} and \texttt{Junyi15} only assign a single KC identifier per question, \texttt{Ednet KT3} and \texttt{Eedi} can assign multiple KC identifiers to the same question. To make a performance prediction for a question our implementation uses a sparse-vector which is 0 everywhere except in the entries which mark the corresponding KC identifiers with a 1 value. 

\subsection*{Total/KC/Question Counts}

Count features summarize a student's history of past interaction with the ITS and are an important component of PFA~\cite{Pavlik2009:Performance} and Best-LR~\cite{Gervet2020:Deep}. In our experiments we evaluate three ways of counting the number of prior correct responses and overall attempts:
\begin{enumerate}
  \item Total counts: Here we compute two features capturing the total number of prior correct responses and overall attempts.
  \item KC counts: For each individual KC we compute two features capturing the number of prior correct responses and attempts related to questions that target the respective KC. A vector containing the counts for all KCs related to the current question is then passed to the machine learning algorithm.
  \item Question counts: Here we compute two features capturing the total number of prior correct responses and attempts on the current question.
\end{enumerate}
All count features are subjected to scaling function $\phi(x) = \log(1 + x)$ before being passed to the machine learning algorithm. This avoids features of large magnitude.

\subsection*{Total/KC/Question Time-Window (TW) Counts}

Time-window based count features summarize student history over different periods of time and provide the model with temporal information~\cite{Choffin2019:DAS3H}. Following the original DAS3H implementation we define a set of time windows $W = \{ 1/24, 1, 7, 30, +\infty \}$ measured in days. For each window $w \in W$, we count the number of prior correct responses and overall attempts of the student which fall into the window. We evaluate three ways of counting the number of prior correct responses and overall attempts:
\begin{enumerate}
  \item Total TW counts: For each time-window, we compute two features capturing the total number of prior correct responses and overall attempts.
  \item KC TW counts: For each time-window and each individual KC we compute two features capturing the number of prior correct responses and attempts related to questions that target the respective KC. A vector containing the counts for all time-windows and KCs related to the current question is then passed to the machine learning algorithm.
  \item Question TW counts: For each time-window, we compute two features capturing the total number of prior correct responses and attempts on the current question.
\end{enumerate}
All count features are subjected to scaling function $\phi(x) = \log(1 + x)$ before being passed to the machine learning algorithm. This avoids features of large magnitude.

\subsection*{R-PFA $F$ \& $R$}

\revinsert{Motivated by the idea that more recently observed student responses are more indicative for future performance than older ones, R-PFA~\cite{Galyardt2015:Move} augments PFA~\cite{Pavlik2009:Performance} by introducing two new features. For each KC $k$ R-PFA considers all interactions of student $s$ with $k$ up to time $t$ and computes: (i) A recency-weighted count of previous failures $F_{s, k, t}$ using exponential decay. (ii) A recency-weighted proportion of past successes $R_{s, k, t}$ using normalized exponential decay. The degree of decay is controlled by the hyperparameters $d_F$ and $d_R \in [0, 1]$. To allow the computation of $R_{s, k, t}$ when a student visits a KC $k$ for the first time, their interaction history is appended with $g = 3$ incorrect ``ghost attempt''. The total number of responses of student $s$ related to KC $k$ is $a_{s,k}$ and correctness indicator $a_{{s, k, i}}$ is 1 when $s$'s $i$-th attempt on KC $k$ was correct and 0 otherwise. With this the two R-PFA features are defined as}
\begin{align}
F_{s, k, t} = \sum_{i = 1}^{a_{s,k}} d_F^{(a_{s,k} + 1) - i} (1 - a_{{s, k, i}})&, \quad
R_{s, k, t} = \sum_{i = (1 - g)}^{a_{s,k}} \frac{d_R^{a_{s,k} - i}}{\sum_{j = (1 - g)}^{a_{s,k}} d_R^{a_{s,k} - i}} a_{{s, k, i}}\,.
\end{align}

\subsection*{PPE Count}

\revinsert{In the conext of vocabulary learning, PPE~\cite{Walsh2018:Mechanisms} was proposed to capture the spacing effect~\cite{Cepeda2008:Spacing}. PPE does so by introducing a weighting scheme which considers the delay between individual practice session and by assuming a multiplicative relationship between the number of prior attempts $a_{s,k}$ with a time variable $T_k$. Further, the model uses a learning rate parameter $c$ and three forgetting rate parameters $x$, $b$ and $m$. These four hyperparameters need to be set by the user. Let $\Delta_{s,k,i}$ be the real time passed since student $s$'s $i$-th response to KC $k$. For our feature evaluation define a weighted count feature $B_{s, k} = a_{s,k}^c T_k^{-d_t}$ by using PPE's multiplicative relationship between $a_{s,k}$ and $T_k$ which are defined as}
\begin{align}
T_k = \left( \sum_{i = 1}^{a_{s,k}} \Delta_{s,k,i}^{1 - x} \right) \left( \sum_{i = 1}^{a_{s,k}} \frac{1}{\Delta_{s,k,j}^{-x}} \right), \quad
d_t = b + m \left( \frac{1}{a_{s,k}} \sum_{i = 1}^{a_{s,k}} \frac{1}{\ln(\Delta_{s,k,i} - \Delta_{s,k,i + 1} + e)} \right).
\end{align}

\subsection*{Current/Prior Elapsed Time}

 Elapsed time measures the time span from question display to response submission~\cite{Shin2021:Saint+}. The idea is that a faster response is correlated with student proficiency. For our experiments we subject elapsed time values to scaling function $\phi(x) = \log(1 + x)$ and also use the categorical one-hot encodings from~\cite{Shin2021:Saint+}. There, elapsed time is capped off at 300 seconds and categorized based on the integer second. We evaluate two variations of this feature. In the first version we compute elapsed time based on interactions with the \emph{current} question. In the second version we compute elapsed time based on interactions with the \emph{prior} question. Because it is unknown how long a student will take to answer a question ahead of time the elapsed time value of the current question is not available for question scheduling purposes and is excluded from the model comparison in Subsection~\ref{subsec:integrating_features}.

\subsection*{Current/Prior Lag Time}

Lag time measures the time passed between the completion of the previous exercise until the next question is received~\cite{Shin2021:Saint+}. Lag time can be indicative for short-term memorization and forgetting. For our experiments we subject lag time values to scaling function $\phi(x) = \log(1 + x)$ and also use the categorical one-hot encodings from \cite{Shin2021:Saint+}. There, lag time is rounded to integer minutes and assigned to one of 150 categories ($0, 1, 2, 3 ,4, 5, 10, 20, 30, \dots, 1440$). Because we cannot compute a lag time value for the very first question a student encounters we use an additional indicator flag. We evaluate two variations of this feature. In the first version we compute lag time based on interactions with the \emph{current} question. In the second version we compute lag time based on interactions with the \emph{prior} question.

\subsection*{Date-Time: Month, Week, Day, Hour}

Date-time features provide information related to the temporal context a learning activity is placed in. Here we consider the month, week, day and hour of interaction. We pass each of these four attribute to the algorithm using one-hot encodings.  

\subsection*{Study Module: One-Hot/Counts}

All four datasets group study activities into distinct categories or modules (e.g. pre-test, effective learning, review, \dots). Providing a machine learning model with information about the corresponding study module can help adapting the predictions to the different learning contexts. \texttt{ElemMath2021} indicates different modules with the \textit{s\_module} attribute. \texttt{EdNet KT3} indicates different modules with the \textit{source} attribute. \texttt{Eedi} indicates different modules with the \textit{SchemeOfWorkId} attribute. For \texttt{Junyi15} we can derive 8 study module identifiers corresponding to the unique combinations of \textit{topic\_mode}, \textit{review\_mode} and \textit{suggested} flags. We encode these study module attribute values into one-hot vectors before passing them to the algorithm.

\subsection*{Teacher/Group ID}

The \texttt{ElemMath2021} dataset annotates each student response with the identifier of the supervising teacher. Similarly, the \texttt{Eedi} dataset annotates each student response with their group identifier. Both attributes provide information about the current learning context. We encode the identifiers into one-hot to allow the machine learning algorithm to learn teacher/group specific parameters.

\subsection*{School ID}

The \texttt{ElemMath2021} dataset associates each student with a physical or virtual tutoring center. Each tutoring center is assigned a unique school identifier. To capture potential differences between the various schools we allow the machine learning algorithm to learn school specific parameters by encoding the school identifiers into one-hot.

\subsection*{Course}

The \texttt{ElemMath2021} dataset contains logs from a variety of different mathematics courses. While each course is assigned a unique course identifier the KCs and questions treated in the individual courses can overlap. To capture differences in the context set by the individual courses we allow the machine learning algorithm to learn course specific parameters by encoding the course identifiers into one-hot.

\subsection*{Topic}

The \texttt{ElemMath2021} dataset organizes learning activities into courses which itself are split into multiple topics. Each topic is assigned a unique topic identifier. The ITS can deploy the same question in multiple topics and the differences in learning context might affect student performance. We allow the machine learning algorithm to learn topic specific parameters by encoding the topic identifiers into one-hot.

\subsection*{Difficulty}

The \texttt{ElemMath2021} dataset associates each question with a manually assigned difficulty score from the set $\{10, 20, 30, 40, 50, 60, 70, 80, 90\}$. We learn a model parameter for each distinct difficulty value by using a one-hot encoding.

\subsection*{Bundle/Quiz ID}

The \texttt{EdNet KT3} datasets annotates each response with a bundle identifier and the \texttt{Eedi} dataset annotates each response with a quiz identifier. Both, bundles and quizzes mark sets of multiple questions which are asked together. The ITS can decide to assign a bundle/quiz to the student which then needs to respond to all associated questions. To capture the learning context provided by the current bundle/quiz we encode the corresponding identifiers into one-hot.

\subsection*{Part/Area one-hot/counts}

The \texttt{Ednet KT3} dataset assigns each question one label based on which of the 7 TOIEC exam parts it addresses. Similarly, the \texttt{Junyi15} dataset assigns each questions one of 9 area identifiers which marks the area of mathematics the question addresses. We allow the machine learning algorithm to learn part/area specific parameters by encoding the part/area identifiers into one-hot. In addition we experiment with two count features capturing the total number of prior correct responses and attempts on questions related to the current part/area. Before passing the count features to the algorithm we subject them to scaling function $\phi(x) = \log(1 + x)$.

\subsection*{Age}

The \texttt{Eedi} dataset provides an attribute which captures students' birth date. We learn a model parameter for each distinct age by using a one-hot encoding. Students without a specified age are assigned a separate parameter.

\subsection*{Gender}

The \texttt{Eedi} dataset categorizes student gender into female, male, other and unspecified. We learn a model parameter for each attribute value by using one-hot vectors of dimension four.

\subsection*{Social Support}

The \texttt{Eedi} dataset provides information on whether students qualify for England’s pupil premium grant (a social support program for disadvantaged students). The attribute categorizes students into qualified, unqualified and unspecified. We learn a model parameter for each attribute value by using one-hot vectors of dimension three.

\subsection*{Platform}

The \texttt{ElemMath2021} dataset contains an attribute which indicates if a question was answered from a physical tutoring centers or the online system. Similarly, the \texttt{EdNet KT3} dataset indicates if a question was answered using the mobile app or a web browser. To pass this information to the model we use a two-dimensional one-hot encoding.

\subsection*{Prerequisite: one-hot/counts}

In addition to a KC model three of the datasets provide a graph structure that captures semantic dependencies between individual KCs and questions. \texttt{ElemMath2021} offers a prerequisite graph that marks relationships between KCs. \texttt{Junyi15} provides a prerequisite graph that describes dependencies between questions. In contrast, \texttt{Eedi} organizes its KCs via a 4-level topic ontology tree. For example the KC \emph{Add and Subtract Vectors} falls under the umbrella of \emph{Basic Vectors} which itself is assigned to \emph{Geometry and Measure} which is connected to the tree root \emph{Mathematics}. To extract prerequisite features from \texttt{Eedi}'s KC ontology we derive a pseudo prerequisite graph by first taking the two lower layers of the ontology tree and then using the parent nodes as prerequisites to the leaf nodes. We evaluate two ways of utilizing prerequisite information for student performance modeling:
\begin{enumerate}
    \item Prerequisite IDs: For each question we employ a sparse vector that is zero everywhere except in the entries that mark the relevant prerequisite KCs (for \texttt{ElemMath2021} and \texttt{Eedi}) or questions (for \texttt{Junyi15}).
    \item Prerequisite counts: For each question we look at its prerequisite KCs (for \texttt{ElemMath2021} and \texttt{Eedi}) or prerequisite questions (for \texttt{Junyi15}). For each prerequisite we then compute two features capturing the number of prior correct responses and attempts related to the respective prerequisite KC or question. After being subjected to scaling function $\phi(x) = \log(1 + x)$, a vector containing the counts for all relevant prerequisite KCs/questions is passed to the machine learning algorithm.
\end{enumerate}

\subsection*{Postrequisite: one-hot/counts}

By inverting the directed edges of the prerequisite graph we derive a postrequisite graph. Analogous to the prerequisite case we encode postrequisite information in two ways: (i) As sparse vectors that are zero everywhere except in the entries that mark the postrequisite KCs/questions with a 1; (ii) As correct and attempt count features computed for each KC/question that is postrequisite to the current question. For further details refer to the above prerequisite feature description.

\subsection*{Video: count/skipped/time}

The \texttt{ElemMath2021} and \texttt{EdNet KT3} dataset both provide information on how students interact with lecture videos. We evaluate three ways of utilizing video consumption behaviour for performance modeling:
\begin{enumerate}
    \item Videos watched count: Here we compute two features: (i) The total number of videos a student has interacted with before; (ii) The number of videos a student has interacted with before related to the KCs of the current question.
    \item Videos skipped count: \texttt{ElemMath2021} captures video skipping events directly. For \texttt{EdNet KT3} we count a video as skipped if the student watches less than 90\%. Again we compute two features: (i) The total number of videos a student has skipped before; (ii) The number of videos a student has skipped before related to the KCs of the current question.
    \item Videos watched time: Here we compute two features: (i) The total time a student has spent watching videos in minutes; (ii) The time a student has spent watching videos related to the KCs of the current question in minutes.
\end{enumerate}
All count and time features are subjected to scaling function $\phi(x) = \log(1 + x)$ before being passed to the machine learning algorithm. This avoids features of large magnitude.

\subsection*{Reading: count/time}

The \texttt{ElemMath2021} and \texttt{EdNet KT3} dataset both provide information on how users interact with reading materials. \texttt{ElemMath2021} captures when a student goes through a question analysis and \texttt{EdNet KT3} creates a log whenever a student enters a written explanation. We evaluate two ways of utilizing reading behaviour for student performance modeling:
\begin{enumerate}
    \item Reading count: Here we compute two features: (i) The total number of reading materials a student has interacted with before; (ii) The number of reading materials a student has interacted with before related to the KCs of the current question.
    \item Reading time: Here we compute two features: (i) The total time a student has spent on reading materials in minutes; (ii) The time a student has spent on reading materials related to the KCs of the current question in minutes.
\end{enumerate}
The count and time features are subjected to scaling function $\phi(x) = \log(1 + x)$ before being passed to the machine learning algorithm. This avoids features of large magnitude.

\subsection*{Hint: count/time}

The \texttt{Junyi15} dataset captures how students make use of hints. Whenever a student answers a question the system logs how many hints were used and how much time was spent on each individual hint. Students are allowed to submit multiple answers to the same question, though a correct response is only registered if it is the first attempt and no hints are used. We evaluate two ways of utilizing hint usage for student performance modeling:
\begin{enumerate}
    \item Hint count: Here we compute two features: (i) The total number of hints a student has used before; (ii) The number of hints a student has used before related to the KCs of the current question.
    \item Hint time: Here we compute two features: (i) The total time a student has spent on hints in minutes; (ii) The time a student has spent on hints related to the KCs of the current question in minutes.
\end{enumerate}
The count and time features are subjected to scaling function $\phi(x) = \log(1 + x)$ before being passed to the machine learning algorithm. This avoids features of large magnitude.

\subsection*{Smoothed Average Correctness}

Let $c_s$ and $a_s$ be the number of prior correct responses and overall attempts of student $s$ respectively. Let $\bar{r}$ be the average correctness rate over all other students in the dataset. The linear logistic model is unable to infer the ratio $c_s / a_s$ of average student correctness on its own. Because of this we introduce the smoothed average correctness feature $r_s$ to capture the average correctness of student $s$ over time as
\begin{equation*}
  \tilde{r}_{s} = \frac{c_s + \eta \bar{r}}{a_s + \eta}.
\end{equation*}
Here, $\eta \in \mathbb{N}$ is a smoothing parameter which biases the estimated average correctness rate, $\bar{r}_s$ of student $s$ towards this all students average $\bar{r}$. The use of smoothing reduces the feature variance during a student's initial interactions with the ITS. A prior work by~\citeNP{Pavlik2020:Logistic} proposed, but did not evaluate, an average correctness feature without the smoothing parameter for student performance modeling. While calibrating this parameter for our experiments, we observed benefits from smoothing and settled on $\eta = 5$.  

\subsection*{Response Pattern}

Inspired by the use of n-gram models in the NLP community (e.g.~\citeNP{Manning1999:Foundations}), we propose \emph{response patterns} as a feature which allows logistic regression models to infer factors impacting short-term student performance. At time $t$, a response pattern $r_t \in \mathbb{R}^{2^n}$ is defined as a one-hot encoded vector that represents a student's sequence of $n \in \mathbb{N}$ most recent responses $w_t = (a_{t - n}, \dots, a_{t - 1})$ formed by binary correctness indicators $a_{t - n}, \dots, a_{t - 1} \in \{0, 1\}$. The encoding process is visualized by Figure~\ref{fig:response_pattern} in the paper.

\section{Appendix: Hyperparameters and Model Specifications}

Here we provide information about the hyperparameter \revinsert{search spaces} used for the deep learning based student performance modeling approaches \revinsert{we} evaluated \revinsert{with a grid search} in  Subsection~\ref{subsec:integrating_features}. For \revinsert{each}\revremove{all} model\revremove{s} we \revinsert{defined a}\revremove{relied on the}  hyperparameter\revremove{s} \revinsert{search space that captures and extends the hyperparameters which were used}\revremove{reported} in the cited references~(DKT, \cite{Piech2015:Deep}; SAKT, \cite{Pandey2019:Self}; SAINT, \cite{Choi2020:Towards}; SAINT+, \cite{Shin2021:Saint+}). A detailed list of the used hyperparameter\revremove{s} \revinsert{spaces} is provided in Table~\ref{tab:hyperparam}. All models were trained for 100 epochs without learning rate decay.

\label{app:hyperparam}
\begin{table}[h]
    \scriptsize
    \centering
    \caption{Hyperparameter\revremove{s} \revinsert{search spaces} for deep learning based approaches.}
    \begin{tabular}{|l|c|c|c|c|}
        \hline
        Model & DKT & SAKT & SAINT & SAINT+ \\\hline
        Hidden \& Embedding Size & $\{50, 100, 200, 500\}$ & $\{50, 100, 200, 500\}$ & $\{64, 128, 256, 512\}$ & $\{64, 128, 256, 512\}$\\
        Number of Layers & $\{1, 2\}$ & $\{1, 2\}$ & $\{2, 4, 6\}$ & $\{2, 4, 6\}$\\
        Dropout Rate & $\{0, 0.2, 0.5\}$ & $\{0, 0.2, 0.5\}$ & - & -\\
        Truncated Sequence Length & - & 200 & 100 & 100\\
        Number of Heads & - & 5 & 8 & 8\\
        Learning Rate & $1 \times 10^{-3}$ & $1 \times 10^{-3}$ & $1 \times 10^{-3}$ & $1 \times 10^{-3}$\\
        Batch Size & $\{128, 256\}$ & $\{128, 256\}$ & $\{128, 256\}$ & $\{128, 256\}$\\
        \hline
    \end{tabular}
    \label{tab:hyperparam}
\end{table}